\pdfoutput=1

\documentclass[11pt]{article}

\usepackage[preprint]{acl}
\usepackage{times}
\usepackage{latexsym}

\usepackage[T1]{fontenc}

\usepackage[utf8]{inputenc}

\usepackage{microtype}

\usepackage{inconsolata}

\usepackage{graphicx}

\usepackage{microtype}
\usepackage{subcaption}
\usepackage{booktabs}
\usepackage{lipsum}
\usepackage{amsfonts}
\usepackage{multirow}
\usepackage{makecell}
\usepackage{epstopdf}
\usepackage{algorithm}
\usepackage{algorithmic}

\usepackage{xcolor}

\title{Exploring and Controlling Diversity in LLM-Agent Conversation}

\author{KuanChao Chu\quad\quad Yi-Pei Chen\quad\quad Hideki Nakayama \\
         The University of Tokyo\\
         \texttt{\{kcchu,ypc\}@nlab.ci.i.u-tokyo.ac.jp}}

\begin{document}
\maketitle
\begin{abstract}
Controlling diversity in LLM-agent simulations is essential for balancing stability in structured tasks with variability in open-ended interactions. However, we observe that dialogue diversity tends to degrade over long-term simulations. To explore the role of prompt design in this phenomenon, we modularized the utterance generation prompt and found that reducing contextual information leads to more diverse outputs. Based on this insight, we propose Adaptive Prompt Pruning (APP), a novel method that allows users to control diversity via a single parameter, $\lambda$. APP dynamically prunes prompt segments based on attention scores and is compatible with existing diversity control methods. We demonstrate that APP effectively modulates diversity through extensive experiments and propose a method to balance the control trade-offs. Our analysis reveals that all prompt components impose constraints on diversity, with the Memory being the most influential. Additionally, high-attention contents consistently suppress output diversity. 
\end{abstract}

\section{Introduction}
\begin{figure}[!t]
    \centering
    \includegraphics[width=\linewidth]{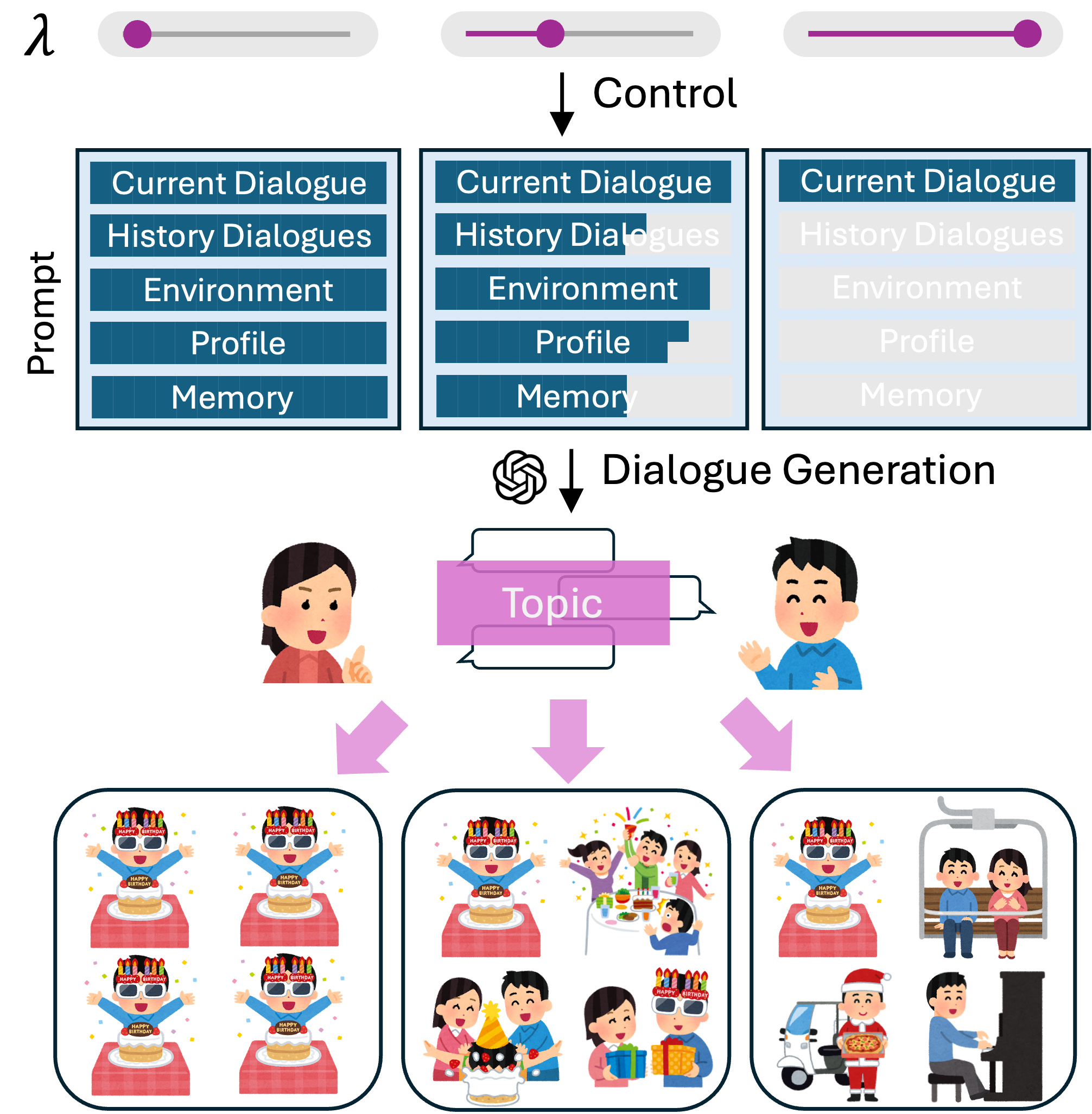}
    \caption{Diversity control in LLM-agent conversations. 
    By increasing $\lambda$, more components are removed from the prompt, selected by their attention scores, thereby enhancing the diversity of the dialogue content.
    }
    \label{fig:overview}
\end{figure}

\begin{figure}
    \centering
    \includegraphics[width=1.0\linewidth]{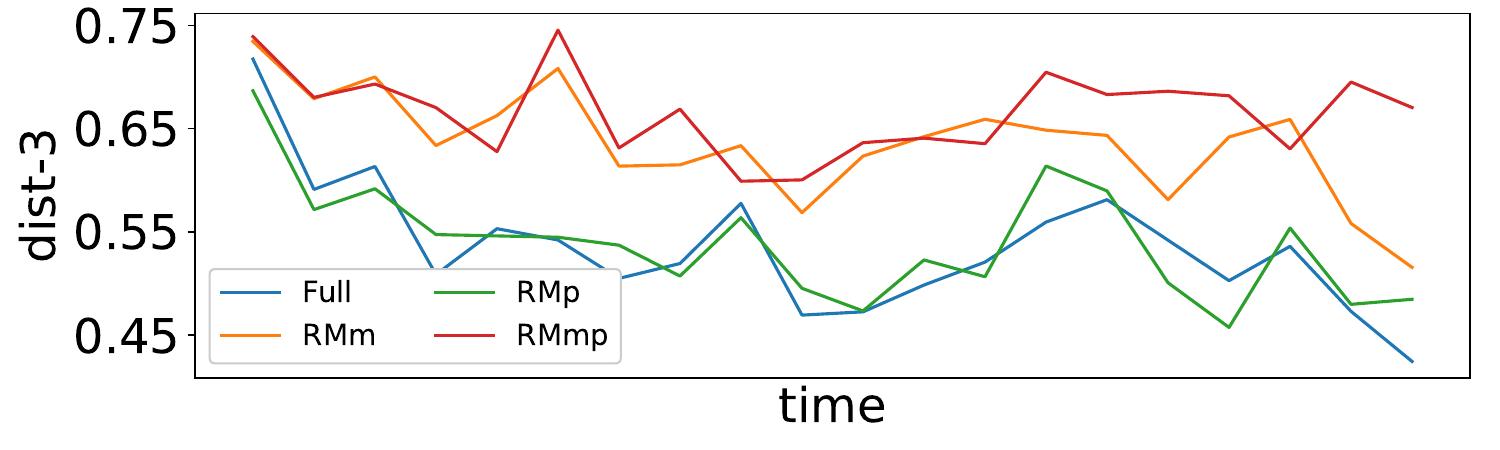}
    \caption{Diversity decreases over time when using the full prompt (Full). Removing memory and previous dialogues from the prompt (RMmp) alleviates this issue.}
    \label{fig:div_vs_time}
\end{figure}

LLM-agent simulation \cite{generative_agent23} can be used across various domains, such as social sciences \cite{zhou2024sotopia,park24_ganext} 
and game development \cite{replica}, where agents interact dynamically to model real-world scenarios. 
By controlling the diversity of dialogues between LLM agents, we can ensure the interactions align with the intended objectives of the simulation.
For instance, in social science research, managing dialogue diversity helps explore how individuals or groups might react, providing insights into human behavior and social dynamics. In game development, a more consistent, static conversation may be preferred for advancing the main storyline, while more varied and creative dialogues can enhance immersion and offer players a unique, personalized experience. 
Meanwhile, as shown in Fig.~\ref{fig:div_vs_time}, we observe a significant decline in dialogue diversity over time, emphasizing the importance of controlling and enhancing diversity in prolonging multi-agent simulation.

In this work, we define diversity as the range of variations generated under identical initial conditions, with a specific focus on LLM-agent conversations. 
The prompt for this task typically comprises several key components: an environment description, agent's profile and memory, dialogue history, and the current dialogue. 
Although most previous works integrate these components into the prompt, it is unclear how these components affect diversity. Does reducing the provided information lead to generalized and less diverse responses, or does it encourage more open and varied outputs? 
Although previous studies have explored the influence of communication structures, the impact of \textit{communication content} on interaction quality remains underexplored \cite{ijcai2024p890}. 

Traditional diversity control methods, such as temperature \cite{ackley1985learning} or top-p sampling \cite{Holtzman2020The}, primarily regulate word distribution during decoding but fail to account for subtle shifts in the input prompt over time. Moreover, these methods can prioritize rare word selection and lead to incoherent output rather than sustaining meaningful diversity. 
On the other hand, prompting each LLM agent to directly generate diverse outputs through sequential generation \cite{treeofthought_23} can be challenging, as it may be harder to follow instructions, and the outcomes often lack contextual coherence. 

To address this gap, we propose Adaptive Prompt Pruning (APP), a removal-based approach for controlling diversity by dynamically adjusting the prompt content via a single parameter, $\lambda$. 
We organized the prompt into blocks, each containing one or more items. Leveraging attention weights from raw output utterances, APP selectively removes items from the modularized prompt. A higher $\lambda$ corresponds to more aggressive pruning and, consequently, a greater potential for diversity. 
We investigate various design choices for the pruning selection, and comprehensively analyze the relation between prompt content and output diversity.

Using data from \citet{generative_agent23} and \citet{humanoid_23}, 
we demonstrate that APP effectively modulates the degree of diversity by pruning influential prompt components. 
Our findings reveal that all prompt components constrain diversity to some extent, with the Memory block having the most significant impact. 
In addition, APP is compatible with established diversity control techniques. 
While increasing diversity through prompt pruning can result in inconsistencies with omitted information, we mitigate this issue by introducing a correction step post-dialogue generation. Experimental results show that this approach balances the trade-off between enhancing diversity and preserving information consistency. 

Beyond pruning, we investigate the role of prompt structure, including the order and length of components, in influencing diversity. Our results indicate that component order significantly affects diversity. Moreover, we analyze the role of pre-existing knowledge within LLMs and its interaction with diversity by replacing agents' names with well-known or rare ones. 

In summary, this paper tackles three fundamental questions related to diversity in multi-agent simulation: 
(1) How can diversity be effectively controlled in multi-agent communication?
(2) How does prompt content influence the level of conversational diversity?  
(3) What trade-offs arise in diversity control, and how can they be mitigated? 
By addressing these questions, we aim to lay the groundwork for understanding and engineering diversity in LLM-based multi-agent systems.

\section{Data, Model, and Task for Diversity Evaluation}

\begin{table}[!th]
\centering
\small
\begin{tabular}{lccc}
\toprule
    \textbf{Block} & \textbf{Item} & \textbf{Word} & \textbf{Type} \\
    \midrule
    Basic Info   
        & 5 & 71.5 & Fixed \\
    Human Needs*   
        & 2\textasciitilde 6 & 20.4 & Fixed in dial.\\
    Memory 
        & 30\textasciitilde 45 & 1318.8 & Trajectory \\
    Previous Dialogues 
        & 1\textasciitilde 3 & 327.4 & Trajectory \\
    Environment 
        & 2 & 69.5 & Context \\
    Current Dialogue   
        & 1 & 284.3 & Context \\
    \bottomrule
\end{tabular}
\caption{The statistics of modularized blocks in the utterance generation prompt, each containing one or more items. *Only appears in the \texttt{HA} dataset. 
}
\label{tab:dataset}
\end{table}

\paragraph{Data}
We leveraged the simulation logs released by Generative Agents \cite{generative_agent23} as our primary dataset, referred to as \texttt{GA}. The logs consist of 290 dialogues simulating agent interactions in a day, which we treated as independent cases. From these, we evenly sampled 20 cases in chronological order for generation and found in preliminary experiments that this quantity was sufficient to align with overall trends. In a conversation, each utterance generated by an LLM agent involves several dynamic steps simulating the internal cognitive behaviors, such as querying related memories, verifying the current environmental states, and integrating these pieces of information into a prompt to produce the final response. For each case, we extracted all necessary contextual information from the logs for accurate simulations, including memory bases, location context, and dialogue history.

We also utilized an extended dataset based on Humanoid Agents \cite{humanoid_23}, \texttt{HA}, which extends \texttt{GA} by introducing new agent states such as basic needs, emotions, and relationship closeness. Following the same methodology, we augmented \texttt{GA}'s 20 cases with these states, collectively termed human needs. Together, these two datasets cover key components of LLM agents and simulation content for human-like behavior \cite{agent_survey_rise_23,agent_survey_exp_24,sumers2024cognitive}.

To better manipulate the prompt for response generation, we modularized \texttt{GA}'s template. We treated the prompt as a sequence of distinct blocks, each comprising multiple units. A unit is the smallest element, either a piece of information (an ``item'', e.g., a single memory string) or an instruction (a ``text'', e.g., ``Here is the memory that is in Eddy Lin's head:''). Table~\ref{tab:dataset} summarizes block specifications. For detailed dataset information, please see the appendix or original papers.

\paragraph{Model}
We used LLaMA 3 and LLaMA 3.1 \cite{llama3_24,llama31_huggingface} as backbone LLMs. 
For practical use, we employed the 8B-Instruct models in half precision. See results on additional models in Appendix~\ref{sec:appendix_other_llm}.

\paragraph{Task}
\textbf{We define diversity as the variation between dialogues generated under identical initial conditions across trials.} In other words, it measures how different dialogues are 
when simulating the same set of LLM agents from the same simulation checkpoint (states).  
For each case, we ran $n=10$ simulations and measured diversity among these $n$ dialogues.
We employed two metrics: similarity (Sim) and distinct-N (Dist-N), which quantify diversity lexically and semantically. The former computes the mean pairwise cosine similarity of dialogue embeddings \cite{reimers2019sentence,wang2021minilmv2}, while the latter measures the proportion of unique N-grams across all $n$ dialogues \cite{li-etal-2016-diversity}. We report the average scores across all cases.

In this paper, the results are mainly from \texttt{GA} on LLaMA 3 unless otherwise specified.

\begin{figure*}[t]
  \centering
    \begin{subfigure}{0.325\linewidth} 
        \includegraphics[width=\linewidth]{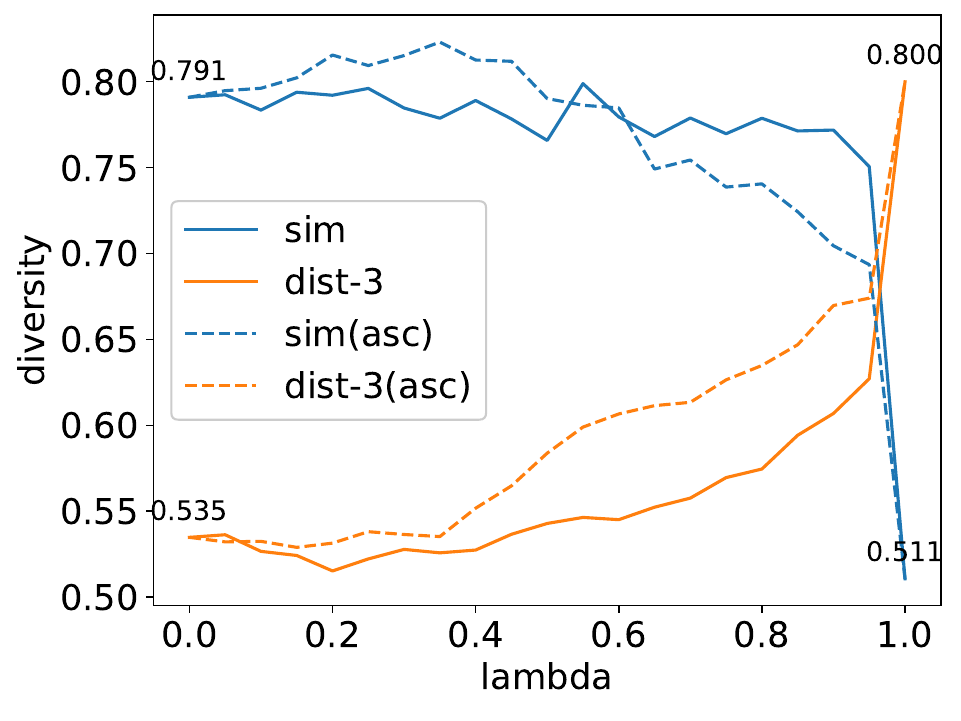}
        \caption{$\lambda$ vs. diversity}
        \label{fig:rmatt_dial}
    \end{subfigure}
    \begin{subfigure}{0.325\linewidth}
        \includegraphics[width=\linewidth]{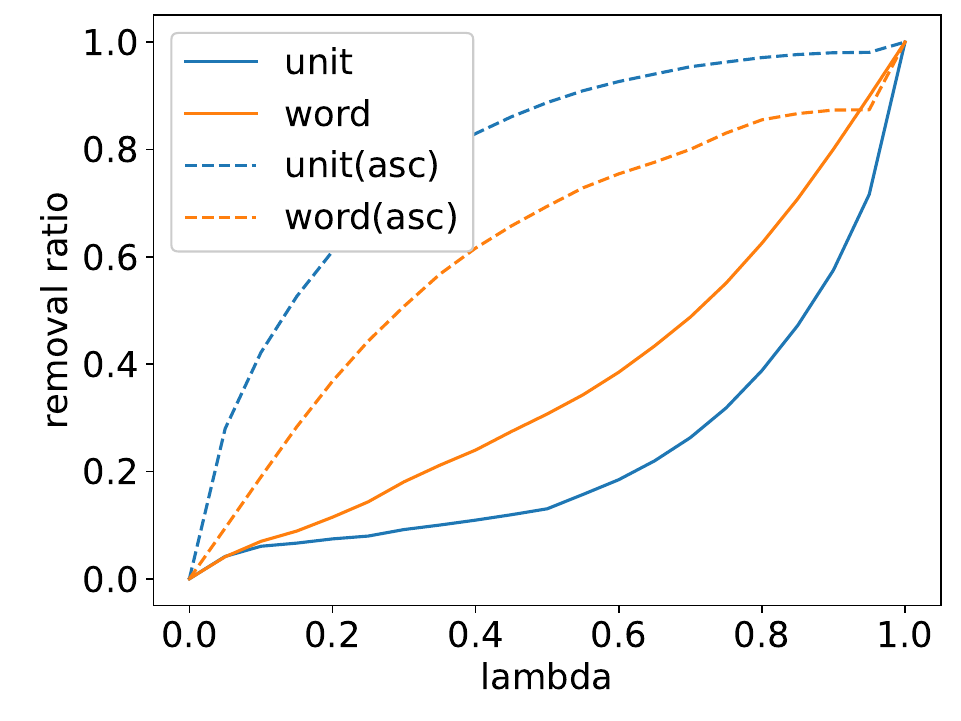}
        \caption{$\lambda$ vs. removal ratio}
        \label{fig:rmatt_ratio}
    \end{subfigure}
    \begin{subfigure}{0.325\linewidth}
        \includegraphics[width=\linewidth]{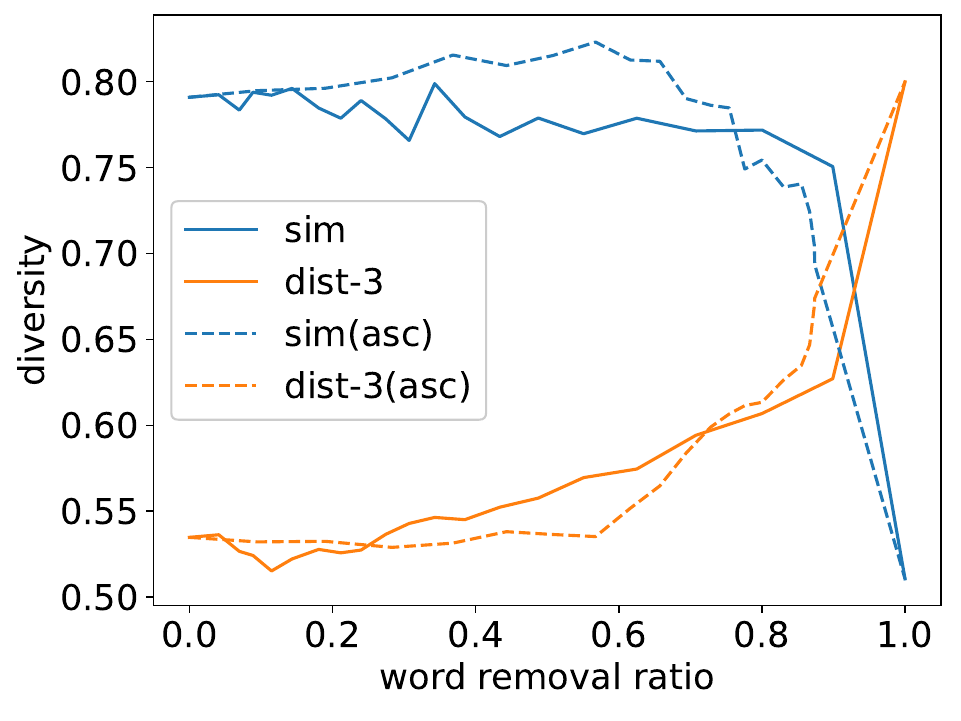}
        \caption{Word removal ratio vs. diversity}
        \label{fig:rmatt_auc} 
    \end{subfigure}
  \caption{Dialogue diversity under our control parameter $\lambda$. As $\lambda$ increases from 0 to 1, diversity generally increases. Removing units based on attention scores in descending order (default) is more word-efficient than removing them in ascending order (asc). Annotated numbers in (a) represent diversity at the endpoints.}
  \label{fig:rmatt}
\end{figure*}

\section{Adaptive Prompt Pruning}
While longer prompts can provide more contextual cues \cite{system2_23} and enrich outputs, they may also constrain generation by imposing stronger priors, leading to more deterministic outputs.

To investigate this trade-off, we conducted a preliminary ablation study by selectively pruning different blocks from the prompt during utterance generation. The resulting changes in diversity are summarized in Table~\ref{tab:rmx}.  
We observed that the degree of diversity varied depending on different block removals. The most notable diversity increase occurred when all four blocks were pruned (RMbmpe), leaving only task instructions and the current dialogue. This indicates that agent-specific information -- particularly memory and environment -- exerts a constraining effect in multi-agent simulations.

Motivated by these findings, we aim to develop a more fine-grained method to control diversity via a single parameter. Specifically, we propose using attention scores to guide the removal of overemphasized prompt segments. This approach adjusts contextual influence without modifying the attention mechanism itself, thus preserving the model's general capabilities. Moreover, it operates independently of specific prompt structures, enhancing its flexibility and applicability. 
Further analysis of content removal is provided in Appendix~\ref{appendix:analysis_removal}.

\begin{table}[!thp]
\centering
\resizebox{0.95\linewidth}{!}{
\begin{tabular}{lcccc}
\toprule

& \textbf{Sim} ($\downarrow$) & \textbf{Dist-1} & \textbf{Dist-2} & \textbf{Dist-3} ($\uparrow$) \\ 
\midrule
Full    & 0.791 & 0.095 & 0.350 & 0.535 \\
\midrule
RMb     & 0.806 & 0.091 & 0.335 & 0.513 \\
RMm     & 0.736 & 0.119 & 0.429 & 0.636 \\
RMp     & 0.802 & 0.095 & 0.352 & 0.538 \\
RMe     & 0.764 & 0.091 & 0.326 & 0.497 \\
RMbmpe  & \textbf{0.511} & \textbf{0.202} & \textbf{0.610} & \textbf{0.800} \\
\bottomrule
\end{tabular}
}
\caption{Diversity changes as blocks are removed from the prompt. RMx represents removing block x, where x corresponds to the initials of the blocks listed in Table~\ref{tab:dataset}.}
\label{tab:rmx}
\end{table}

\subsection{Method}

We compute attention-based importance scores for each prompt unit based on the generated response. Given a full prompt as input, the model generates an output sequence $r = \{t_{r_{1}},t_{r_{2}},...,t_{r_{n}}\}$. 
Each unit $u$ in the prompt is defined as a sequence of tokens $u = \{t_{u_{1}},t_{u_{2}},...,t_{u_{m}}\}$. The attention values from $r$ to $u$ can be represented by a tensor $a \in \mathbb{R}^{L \times H \times m \times n}$, where $L$ is the number of attention layers and $H$ is the number of attention heads. 
To facilitate comparison between units, we compress $a$ to $a' \in \mathbb{R}^{L \times H}$ using a Reducer function $a' = R(a)$. Specifically, we apply a \textit{sum-mean} reducer\footnote{See Appendix~\ref{appendix:analysis_removal} for further analysis on Reducer design choices and their effects.}: summing over the $m$ dimension (tokens in $u$) and averaging over the $n$ dimension (tokens in $r$). This captures the overall influence of unit $u$ on the generated response.
We then aggregate $a'$ into a scalar attention score $a_u$ by averaging over heads and summing across layers:
\begin{equation}
    a_u = \sum_{i=1}^{L} \frac{1}{H} \sum_{j=1}^{H} a'_{i,j}
\end{equation}

\begin{algorithm}[ht]
\caption{Attention-based Unit Removal}
\label{alg:rmatt}
\textbf{Input}: Units $U$, Scores $\{a_u\}$, Removal factor $\lambda$\par
\begin{algorithmic}[1]
\STATE Sort removable units $U_{rm} \subseteq U$ by $a_u$ in descending order
\STATE Set $S_{target} = \lambda \cdot \sum_{u \in U_{rm}} a_u$, $current\_sum \gets 0$, $U_{rm}' \gets \emptyset$

\STATE \textbf{for} each $u \in U_{rm}$ \textbf{do}
\STATE \hspace{1em} \textbf{if} $current\_sum + a_u \leq S_{target}$ \textbf{then}
\STATE \hspace{2em} $current\_sum \gets current\_sum + a_u$
\STATE \hspace{2em} Add $u$ to $U_{rm}'$
\STATE \hspace{1em} \textbf{if} $current\_sum \geq S_{target}$ \textbf{then}
\STATE \hspace{2em} \textbf{break}

\STATE Remove $U_{rm}'$ from full prompt
\end{algorithmic}
\end{algorithm}

Next, we introduce a single parameter $\lambda \in [0,1]$ to control the pruning intensity. 
We first define a set of \textit{removable units} $U_{rm}$, excluding essential elements such as task or output instructions. The units in $U_{rm}$ are ranked by their $a_u$ scores in descending order. 
We then select the top-ranked units such that their cumulative score reaches $\lambda$ times the total score of $U_{rm}$.
These selected units are removed from the prompt before generating each utterance.
This process is applied to each utterance generation step. Algorithm~\ref{alg:rmatt} outlines the procedure.

In our implementation, $a_u$ scores are averaged over three sampled responses for robustness. 
$U_{rm}$ consists solely of ``item''-type units except the one from ``Current Dialogue''.
If all items in a block are removed, the whole block is discarded.

\subsection{Discussion}
\label{sec:app_discuss}

\begin{figure*}[t]
  \centering
    \begin{subfigure}{0.325\linewidth}  
        \includegraphics[width=\linewidth]{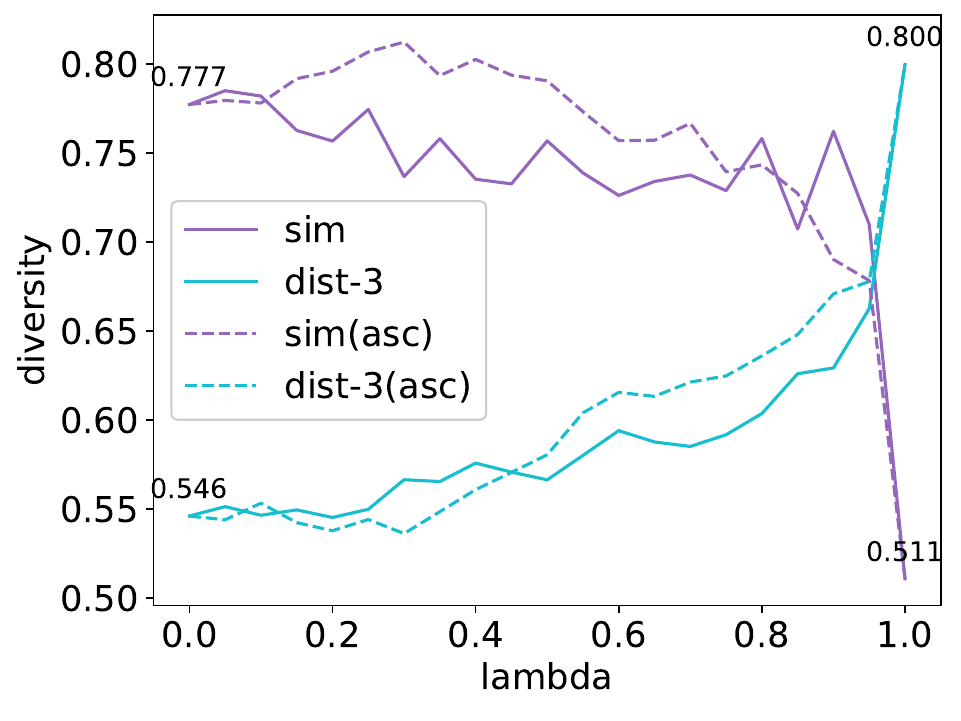}
        \caption{LLaMA 3, \texttt{HA}}
        \label{fig:rmatt_v3}
    \end{subfigure}
    \begin{subfigure}{0.325\linewidth}
        \includegraphics[width=\linewidth]{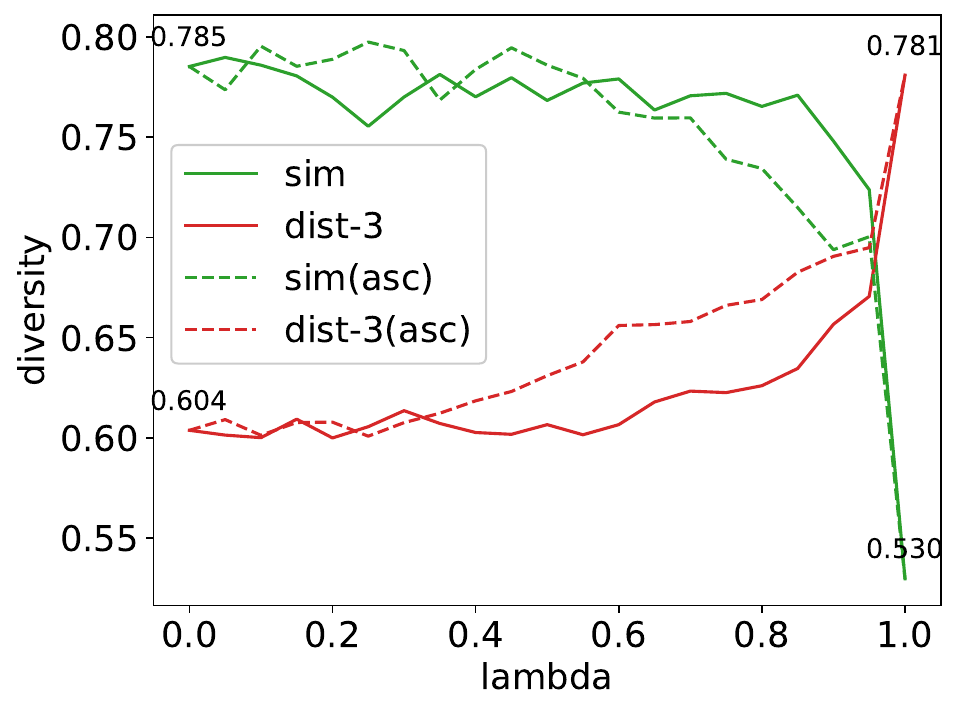}
        \caption{LLaMA 3.1, \texttt{GA}}
        \label{fig:rmatt_31}
    \end{subfigure}
    \begin{subfigure}{0.325\linewidth}
        \includegraphics[width=\linewidth]{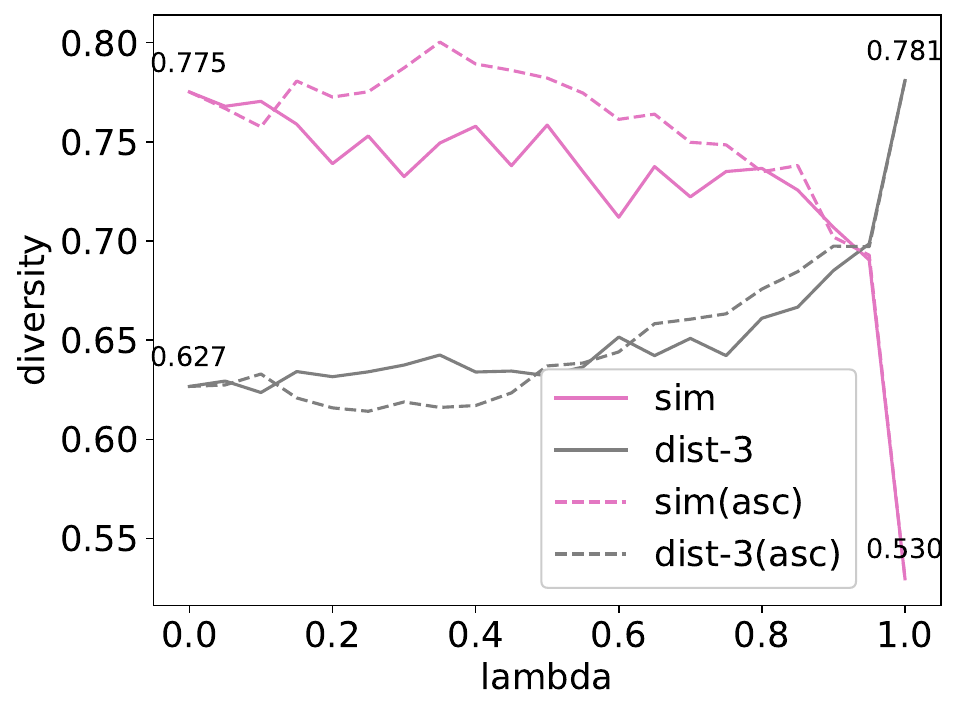}
        \caption{LLaMA 3.1, \texttt{HA}}
        \label{fig:rmatt_31v3}
    \end{subfigure}
  \caption{Results for $\lambda$ vs. diversity under different model and data settings. Similar trends are observed as in the LLaMA 3, \texttt{GA} setting, despite differences in initial diversity. Annotated numbers indicate diversity at the endpoints.}
  \label{fig:rmatt2}
\end{figure*}

\begin{figure*}[t]
  \centering
  \begin{subfigure}[t]{0.325\linewidth}
      \includegraphics[width=\linewidth]{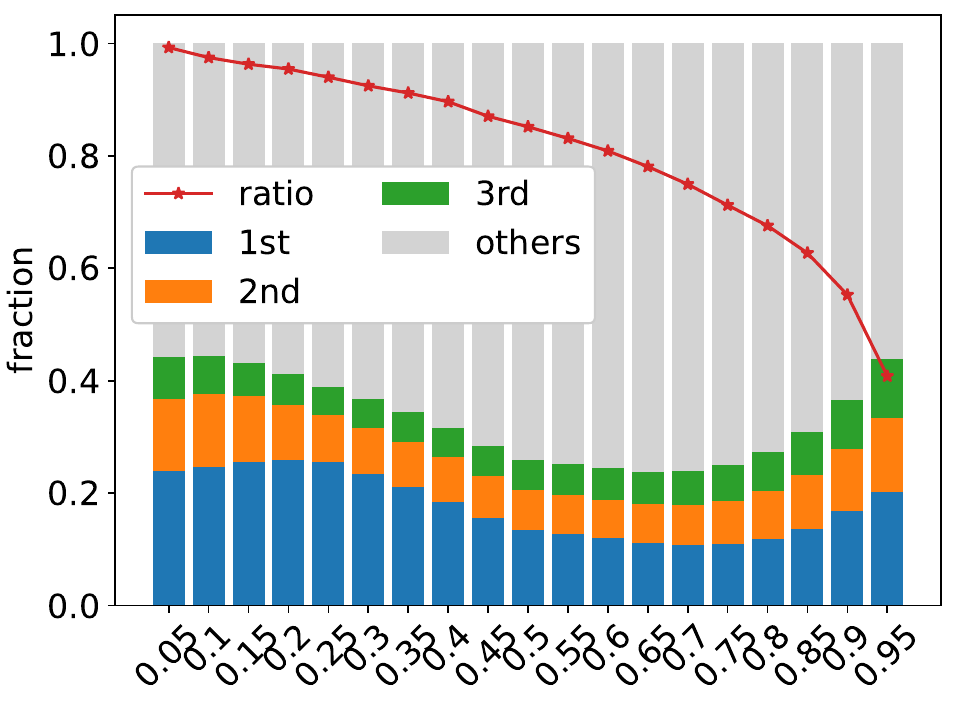}
      \caption{Post-removal stats for different lambda values: Top 3 unit score share (bars) and total score retention (line)}
      \label{fig:post_attn}
  \end{subfigure}
  \hspace{0.1\linewidth}
  \begin{subfigure}[t]{0.325\linewidth}
      \includegraphics[width=\linewidth]{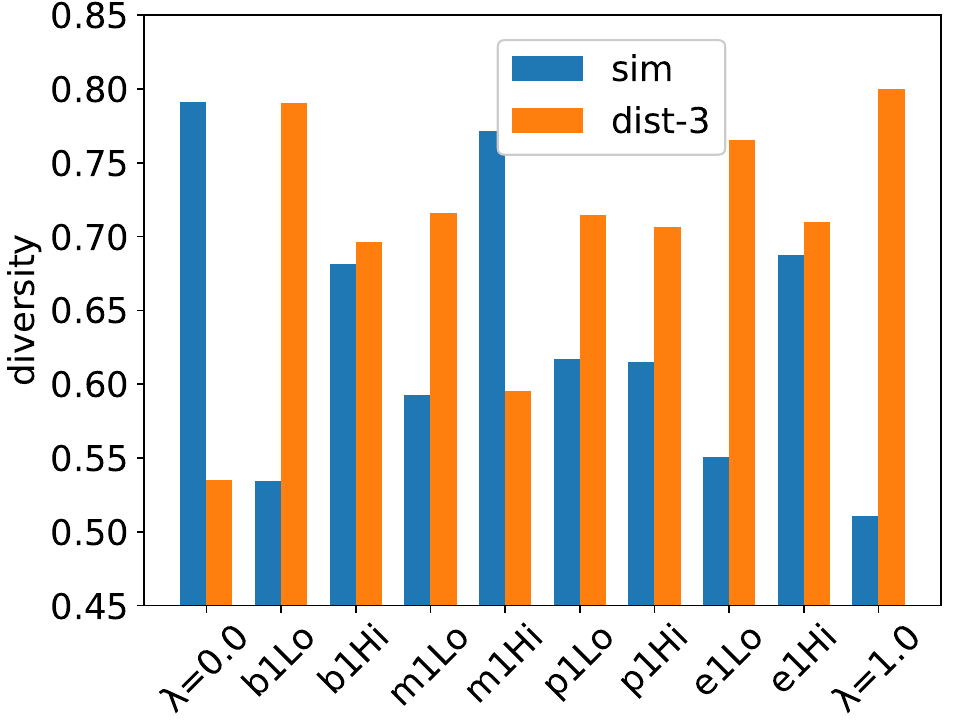}
      \caption{Retain-1: Keep only one removable unit in the prompt, selected from various blocks (`x' in \texttt{x1Hi}/\texttt{x1Lo} denotes the initial letter of a block name)}
      \label{fig:rt1}
  \end{subfigure}
  \caption{More analysis of Adaptive Prompt Pruning discussed in Section~\ref{sec:app_discuss}.}
  \label{fig:rmatt_analysis}
\end{figure*}

\paragraph{Main Results}
We evaluated dialogue diversity across varying $\lambda$ values. Fig.~\ref{fig:rmatt_dial} and Fig.~\ref{fig:rmatt_ratio} show that as $\lambda$ increases, diversity generally rises, confirming that $\lambda$ provides effective control. Since units are removed in descending order of $a_u$, only a small number needed to be pruned at low $\lambda$ values to induce noticeable effects. This observation motivates a new efficiency criterion: methods that achieve higher diversity with fewer content removals are more efficient. Accordingly, we plot diversity as a function of word removal ratio in Fig.~\ref{fig:rmatt_auc}.

To assess our selection strategy, we compared descending-order removal (prioritizing high-attention units) with ascending-order removal (prioritizing low-attention units). Although the ascending strategy sometimes yields higher diversity (especially in the Dist-3 metric), it requires more unit removal for the same $\lambda$. As Fig.~\ref{fig:rmatt_auc} shows, the descending strategy is more efficient overall, except near $\lambda = 1.0$.

We further tested this method across different models and datasets (Fig.~\ref{fig:rmatt2}). Among them, LLaMA-3.1 achieves higher diversity at $\lambda = 0.0$. Similarly, the \texttt{HA} dataset, despite its longer prompts, starts with higher initial diversity (e.g., Dist-3 increases from 0.535 to 0.546 under LLaMA-3), likely due to the inclusion of human needs information that broadens response space. However, as units are pruned, diversity increases across all settings, peaking at $\lambda = 1.0$, suggesting that these additional units function more as constraints than generators of variability.

\paragraph{Post-removal Attention Scores}
While Fig.~\ref{fig:rmatt_dial} and Fig.~\ref{fig:rmatt_auc} show that diversity correlates with $\lambda$ and word removal ratio, the modest diversity gains at low $\lambda$ remain puzzling. To explore this, Fig.\ref{fig:post_attn} reports two post-removal metrics: (1) the total attention scores of the remaining $U_{rm}$ expressed as a percentage of the original scores (red line), and (2) the shares of the top-3 removable units.

Since attention redistributes after removal, $\lambda$ does not translate linearly into reduced attention. For example, at $\lambda = 0.6$, attention drops by only 19\% (red line in Fig.\ref{fig:post_attn}). 
This effect is particularly evident for smaller $\lambda$, which explains the limited growth in diversity during the early stages shown in Fig.~\ref{fig:rmatt}. In this regime, attention on removable units decreases only marginally, while the proportion of the top-1 unit's score even rises. 
Moreover, when $\lambda$ exceeds 0.8, the top units' attention proportion increases, contradicting the trend of growing diversity. 
This phenomenon is likely a consequence of the sharp reduction in the number of remaining removable units.

\paragraph{Retain-1 Analysis}
To isolate the influence of specific items, we conducted a controlled experiment where only one item is retained in the prompt, thereby minimizing confounding effects from attention redistribution. 
As shown in Fig.~\ref{fig:rt1}, retaining the highest-attention item (Hi) generally leads to lower diversity than retaining the lowest-attention item (Lo) in the same block, reaffirming that \textbf{high-attention content tends to suppress diversity}. 

The Hi and Lo settings of each block type exhibit nuanced differences. 
For the \textit{Previous Dialogues} block, the gap between \texttt{p1Lo} and \texttt{p1Hi} is smaller, likely because the block occasionally contains only one item, leaving little room for a Hi/Lo difference.
The \textit{Memory} block, however, has the most detrimental effect on diversity across all blocks. Even a single \textit{Memory} item can substantially reduce diversity (e.g., Dist-3 drops from 0.800 to 0.595 for \texttt{m1Hi}), suggesting that the model treats \textit{Memory} information as particularly constraining.

Interestingly, this also sheds light on the efficiency reversal between sorting strategies observed at the tail of Fig.~\ref{fig:rmatt_auc}. At $\lambda = 0.95$, under descending order, 83.4\% of the remaining items are from \textit{Memory}. In contrast, ascending order keeps mostly \textit{Previous Dialogues} (59.3\%), with \textit{Memory} at only 1.6\%. 
This shift in block composition likely accounts for the ascending strategy’s advantage in the high-$\lambda$ regime.

From a Transformer perspective, attention scores serve as an informative signal of unit influence, pinpointing those that disproportionately shape generation (Fig.~\ref{fig:rt1}). By exploiting this signal, APP improves diversity through targeted pruning of overemphasized segments, strategically redirecting attention away from dominant contextual anchors.

\section{Balancing Diversity Trade-off}
Using unit removal is an effective method to control and enhance dialogue diversity. However, the generated responses may conflict with the pruned information. To address this issue, we introduce an additional step for revision to rectify potential discrepancies in the generated utterances.

\subsection{Method}
\begin{figure}[!th]
\centering
\includegraphics[trim={1.8cm 3cm 3.5cm 4cm},clip,width=1.0\linewidth]{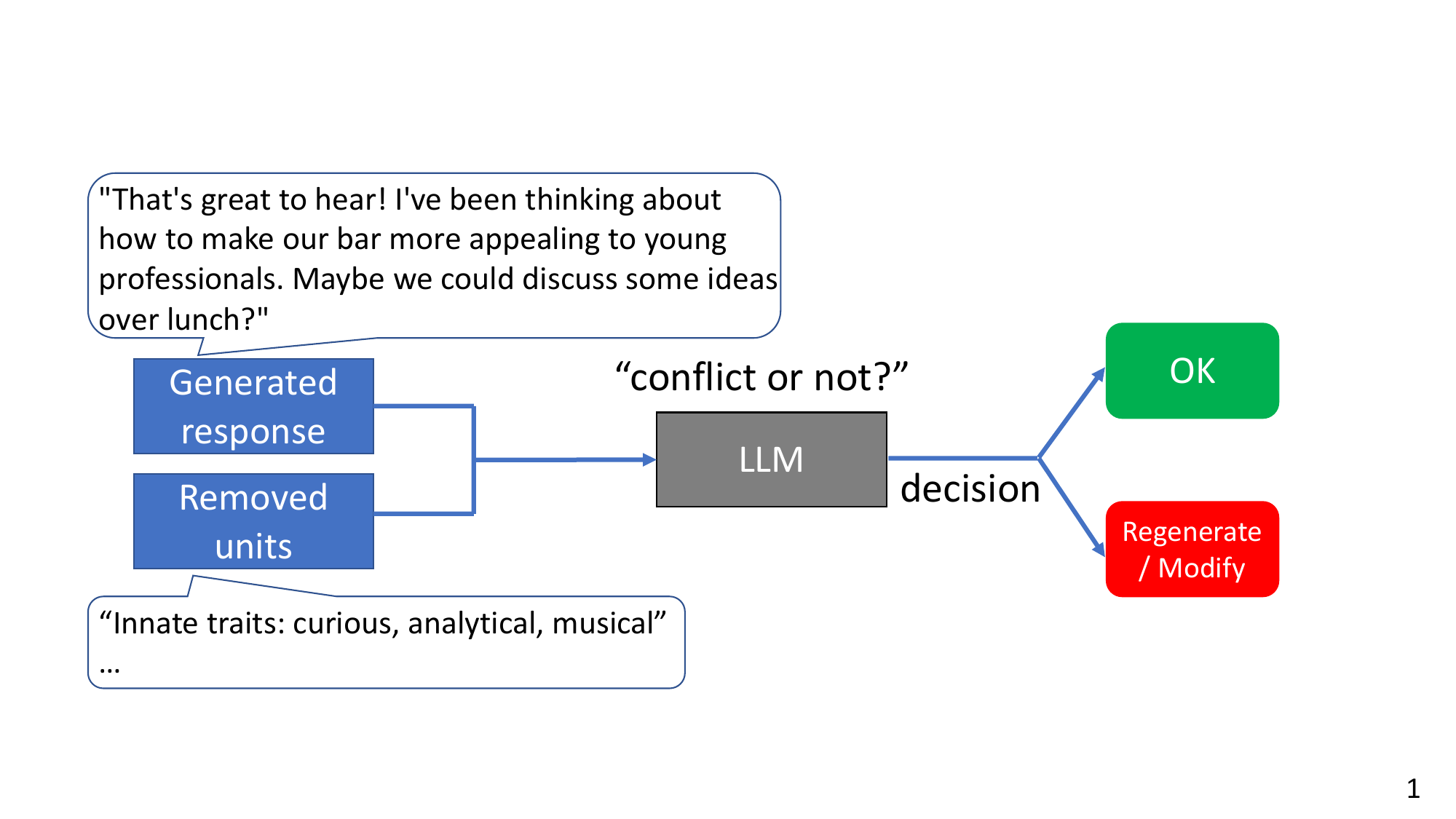}
\caption{An illustrative figure depicting the revision process after generation with a units-removed prompt.}
\label{fig:correction}
\end{figure}

After generating a response controlled by $\lambda$, we collect the removed units and the generated utterance to assess whether the utterance conflicts with the content of the removed units. If a conflict is detected, the utterance undergoes revision; otherwise, it is accepted as is. Fig.~\ref{fig:correction} illustrates this workflow. In our implementation, we use the same LLM for conflict detection, utilizing the following task prompt: ``\{name of agent A\} is now in a chat with \{name of agent B\} and going to say '\{response\}'. Are there any inconsistencies between this response and the statements above?'' The LLM generates a comment and assigns a score from 1 to 10, where higher scores indicate greater inconsistency\footnote{We did not separately validate LLM assessments, but relied on established evidence \cite{zheng2023judging,thakur2024judging} showing LLM-as-Judge aligns well with human judgment, as our focus is on diversity control.}. We take the average of three scoring runs as the final score and set a threshold $\theta = 6.67$. If the score exceeds $\theta$, a conflict is identified.
When a conflict occurs, there are two common revision approaches: 
(1) Regenerating: Revert to the previous stage to generate a new response.
(2) Comment-based modification: Revise the utterance based on the generated comments \cite{pan2023automaticallycorrectinglargelanguage}. 
For simplicity, this study adopts the first approach by preparing multiple backup responses during the initial generation. The rollback process is repeated up to three times until the score drops below $\theta$, or the utterance with the lowest score is selected.

\begin{figure}[!th]
  \centering
    \begin{subfigure}{0.49\linewidth}  
        \includegraphics[width=\linewidth]{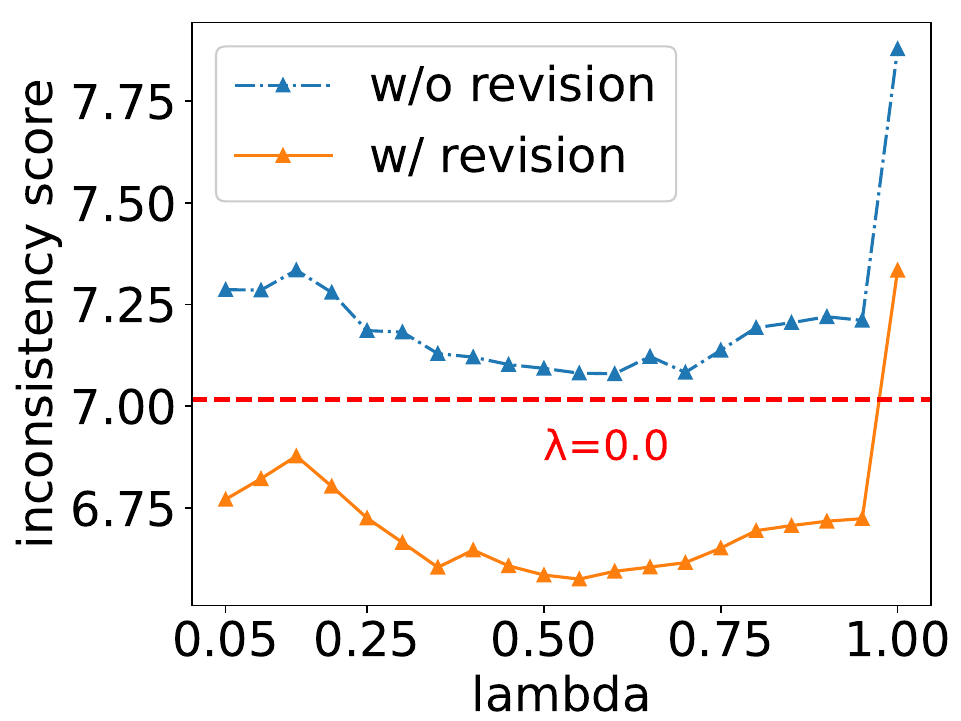}
        \caption{$\lambda$ vs. inconsistency score} 
        \label{fig:corr_evl}
    \end{subfigure}
    \begin{subfigure}{0.49\linewidth}
        \includegraphics[width=\linewidth]{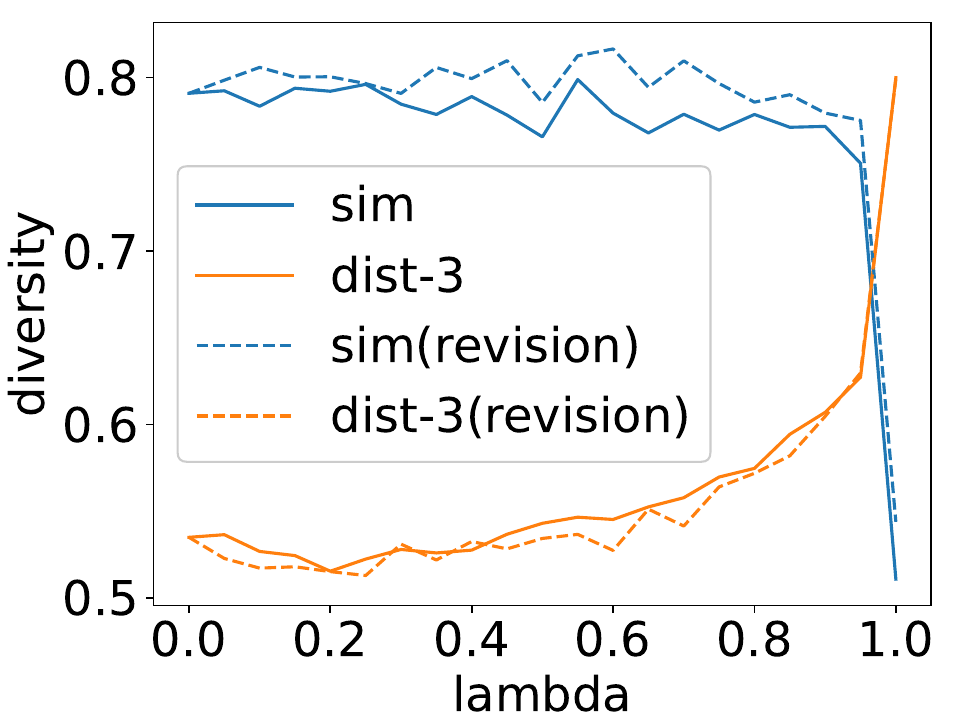}
        \caption{$\lambda$ vs. diversity} 
        \label{fig:corr_div}
    \end{subfigure}
  \caption{Comparison of results with and without revision. $\lambda=0.0$ is a special case without pruning and revision. Its inconsistency score is estimated and illustrated as the red line in (a). 
  }
  \label{fig:corr_res}
\end{figure}

\subsection{Discussion}
Fig.~\ref{fig:corr_evl} compares the average inconsistency scores of dialogues before and after applying revision. 
As a baseline, we also estimate the score for $\lambda=0.0$, which does not involve unit removal but instead uses the same task prompt to assess consistency between the content of all units in the full prompt and the response. 
The results indicate that $\lambda=0.0$ and $\lambda=1.0$ correspond to the lowest and highest inconsistency scores without revision, respectively. 
However, the correlation between the degree of removal and the inconsistency score is not straightforward (e.g., the second-highest score occurs for $\lambda=0.15$, where fewer words are removed compared to higher $\lambda$ values). 
This may be because the error space for open-ended conversations is smaller than that for task-oriented ones, making larger $\lambda$ values unnecessary for introducing errors. 

\begin{table}[ht]
\centering
\resizebox{1.0\linewidth}{!}{
\begin{tabular}{llcccccc}
\toprule
\textbf{Config} & \textbf{} & \textbf{Sim ($\downarrow$)} & \textbf{Dist-3} & \textbf{Flu.} & \textbf{Coh.} & \textbf{Overall} & \textbf{Len} \\
\midrule
\multirow{2}{*}{Default}     
& Full & .791 & .535 & 4.9 & 4.1 & 4.1 & 39.9 \\
& APP  & \textbf{.771} & \textbf{.594} & \textbf{5.0} & \textbf{4.5} & \textbf{4.5} & 38.4 \\
\midrule
\multirow{2}{*}{T = 1.0}       
& Full & .791 & .578 & 4.9 & 4.2 & 4.3 & 40.1 \\
& APP  & \textbf{.778} & \textbf{.634} & \textbf{5.0} & \textbf{4.6} & \textbf{4.6} & 38.7 \\
\midrule
\multirow{2}{*}{p = 0.99}      
& Full & .800 & .569 & \textbf{5.0} & 4.3 & 4.3 & 40.0 \\
& APP  & \textbf{.776} & \textbf{.624} & \textbf{5.0} & \textbf{4.5} & \textbf{4.6} & 38.4 \\
\midrule
\multirow{2}{*}{Sequential}  
& Full & \textbf{.634} & .695 & 4.9 & 3.5 & 3.6 & 21.9 \\
& APP  & .645 & \textbf{.740} & \textbf{5.0} & \textbf{4.0} & \textbf{4.1} & 21.3 \\
\bottomrule
\end{tabular}
}
\caption{APP is efficient and further enhances diversity with various strategies. The default decoding parameters are T=0.8 and p=0.9, while selecting $\lambda$=0.85 for APP.}
\label{tab:compat}
\end{table}

We also examined potential biases that may arise when an LLM is used to evaluate its own generated content \cite{panickssery2024llm}, and assessed their impact on our method.
We replaced LLaMA 3 with GPT-4o-mini for conflict detection.
The results show that although the scoring ranges differed (GPT-4o-mini: 3.00--6.68; LLaMA 3: 7.02--7.88), both models exhibited similar trends. For example, inconsistency scores were lowest at $\lambda=0.0$  and highest at $\lambda=1.0$. These findings suggest that the current evaluation setup did not substantially bias the overall conclusions.

After revision, the scores consistently reduced, indicating that the model finds the revised responses more faithful. Notably, the revised scores are even lower than those for $\lambda=0.0$, suggesting that the model perceives flaws in outputs generated with the full prompt, which the revision process helps to mitigate.
Regarding diversity, Fig.~\ref{fig:corr_div} shows the diversity metrics with and without revision. While some metrics reveal slight reductions within specific $\lambda$ ranges, the overall results demonstrate that our method effectively enhances diversity while maintaining consistency between the utterance and all units.

\section{Comparing Diversity Approaches}

We compared APP with two common methods for controlling generation diversity: (1) tuning decoding parameters and (2) sequential generation, with results summarized in Table~\ref{tab:compat}.

\textbf{Decoding parameter tuning} (e.g., increasing temperature $T$ or top-$p$) is a widely adopted strategy to boost diversity by encouraging low-probability token selection.  
However, Table~\ref{tab:compat} shows that neither higher $T$ nor $p$ achieved the same diversity gains as APP. Notably, combining APP with these methods further improved results -- for instance, Dist-3 rose from 0.578 to 0.634 when APP was used with $T=1.0$.

\textbf{Sequential generation} produces multiple responses in parallel, conditioning each on the preceding ones to promote topical variation. We implemented this by appending ``Please output TEN candidates'' to the prompt and selecting one at random. This method largely improves diversity, aligning with findings by \citet{treeofthought_23}. However, it often led to shorter, less coherent outputs, and can hinder desired output formatting. 

In addition to diversity evaluation, we also measured fluency, coherence, and the overall dialogue quality by GPT-4 on a 5-point Likert scale, following \citet{mendonca-etal-2024-benchmarking}. Results demonstrate that our adaptive pruning maintains perfect output fluency and even yields higher coherence and overall scores compared to the baseline.

In summary, APP (1) enables direct control of diversity via context pruning, ideal for heterogeneous agent simulation; (2) outperforms decoding-based methods and complements sequential generation without harming coherence; and (3) can be combined with other strategies for further gains.

\section{Other Factors Affecting Diversity}
Beyond using unit removal to control and enhance diversity, we also explored factors influencing diversity in the original prompt, specifically block order, name frequency, and block length. The results are shown in Table~\ref{tab:explore} and Table~\ref{tab:explore_appendix}. 
We also provide more analysis in Appendix~\ref{appendix:analysis_other_factors}. 

\paragraph{Block order critically affects diversity}
The reasoning abilities of LLMs are influenced by premise order \cite{chen2024premise} and critical information placement \cite{liu2024lost}. 
This experiment examined whether input order also impacts dialogue diversity. We rearranged blocks in the prompt (denoted by the sequence of their initials) and observed that sequence substantially influences diversity. 
For instance, reversing \texttt{bpmec} to \texttt{cempb} drastically reduced quality and diversity, with Dist-3 dropping from 0.535 to 0.191. 
Under \texttt{cempb}, the generated dialogue started to repetitively cycle through the same round\footnote{We calculate the duplication rate of the final utterance in a dialogue. In the \texttt{cempb} setting, the rate is 66.5\%, compared to 7.9\% in the original Full results.}, leading to a significant degradation in Dist-N. Notably, the amplified context differences caused by such repetition also reduced sim scores, an embedding-based measure.
A key negative pattern was placing \texttt{c} first and \texttt{b} last. 
Comparing \texttt{bmepc} and \texttt{bmecp} (with Dist-3 scores of 0.514 and 0.413, respectively) revealed that placing \texttt{p} before \texttt{c} mitigates significant drops in diversity. This pattern aligns with chronological order, highlighting the importance of block order in ensuring greater initial diversity.

\paragraph{A frequent name can enhance diversity as parametric knowledge is amplified} 
We used name replacement to analyze the agent's reliance on parametric and in-context knowledge in dialogue generation and its impact on diversity.
Inspired by the input sensitivity hypothesis in \cite{ember_23}, which links input frequency to task performance, we replaced agent names with two sets of fictional characters: ``Harry Potter with Severus Snape (\texttt{HPSS})'' and ``Tifa Lockhart with Cloud Strife (\texttt{TLCS}).'' Names are distinctive nouns that can strongly invoke learned knowledge, making them useful for this analysis.
According to the C4 dataset \cite{c4dataset_21}, a common LLM pretraining corpus, ``Harry Potter'' appears 762,023 times, whereas ``Tifa Lockhart'' appears only 432 times. 
This disparity suggests differing learned strengths, potentially affecting the model's use of parametric knowledge.

Results show that replacing names alone did not improve diversity (\texttt{HPSS} to Full). 
However, when prompts were further pruned (RMbmp), name replacement significantly boosted diversity, as shown by Dist-N (\texttt{HPSS}+RMbmp to RMbmp). 
Comparing name combinations (\texttt{HPSS}+RMbmp to \texttt{TLCS}+RMbmp) revealed that high-frequency names had a stronger effect. 
This suggests that pruning strengthens parametric knowledge, enabling outputs to integrate both parametric and in-context information, enhancing diversity. 
Notably, this manifests larges on additional vocabulary\footnote{e.g., potion, wizard (\texttt{HPSS}); shinra, soldier (\texttt{TLCS})} in dialogue, increasing distinct n-grams with minimal impact on embeddings. 
Overall, the experiment highlights how LLM agents utilize both knowledge sources, shedding light on their interplay and impact on diversity.

\begin{table}[!thp]
\centering
\small
\resizebox{1.0\linewidth}{!}{
\begin{tabular}{lcccc}
\toprule

& \textbf{Sim} ($\downarrow$) & \textbf{Dist-1} & \textbf{Dist-2} & \textbf{Dist-3} ($\uparrow$) \\ 
\midrule
Full       & 0.791 & 0.095 & 0.350 & 0.535 \\
\midrule
\multicolumn{5}{c}{\textbf{Block Order}} \\
\midrule
bpmec & 0.789 & \textbf{0.098} & \textbf{0.352} & \textbf{0.535} \\
bmepc & 0.787 & 0.094 & 0.339 & 0.514 \\
bmecp & 0.761 & 0.081 & 0.276 & 0.413 \\
cepmb & \textbf{0.744} & 0.053 & 0.145 & 0.206 \\
cempb & 0.747 & 0.050 & 0.135 & 0.191 \\
\midrule
\multicolumn{5}{c}{\textbf{Name Frequency}} \\
\midrule
\texttt{HPSS}       & 0.828 & 0.093 & 0.337 & 0.518 \\
RMbmp      & \textbf{0.693} & 0.143 & 0.495 & 0.706 \\
\texttt{TLCS}+RMbmp & 0.733 & 0.143 & 0.501 & 0.713 \\
\texttt{HPSS}+RMbmp & \textbf{0.693} & \textbf{0.176} & \textbf{0.553} & \textbf{0.761} \\

\bottomrule
\end{tabular}
}
\caption{Diversity changes resulting from altering block order and name frequency in the text space.}
\label{tab:explore}
\end{table}

\section{Related Work}
Research in LLM-based multi-agents has explored effective collaboration and meaningful interaction between multiple agents to achieve a predefined goal or to simulate human behavior. 
The former are task-oriented, studying the communication strategy \cite{liu2024a} or the collaboration between agents of different roles such as a program manager and a software engineer for software development \cite{chen2024agentverse,hong2024metagpt}. 
The latter are open-domain, investigating emergent human behavior or social simulation \cite{generative_agent23,gao2024simulating}. 
However, most focus on task performance metrics rather than the intrinsic qualities of agent interactions. \citet{chu2024cohesive} revealed the repetition, inconsistency, and hallucination problems in LLM-based multi-agent conversations. 

Diversity in natural language generation has long been a critical research challenge. 
Techniques such as temperature scaling \cite{ackley1985learning} or nucleus sampling \cite{Holtzman2020The} have been explored to generate varied responses while maintaining coherence. 
\citet{stasaski-etal-2020-diverse}  progressively collects more diverse training data based on a diversity metric. 
To reduce the cost of enhancing diversity, \citet{lee2022factuality} further improves upon nucleus sampling, achieving better trade-offs between generation diversity and factuality. Similarly, \citet{chung-etal-2023-increasing} increases text generation diversity while maintaining data accuracy through human interventions.

Balancing diversity and relevance in multi-turn dialogues remains non-trivial. Studies such as \citet{li-etal-2016-diversity} have investigated diversity-promoting objectives like Maximum Mutual Information (MMI) to address response repetition in dialogue systems. 
\citet{zhou-etal-2023-simoap} generated a large number of utterance candidates and selected the best one using NLI entailment scores to achieve the generation of diverse and coherent dialogues. 
However, controlling diversity in multi-agent conversations is still underdeveloped. 
\citet{chu2024cohesive} applied dynamic similarity threshold to remove repetitive utterances.
Our work bridges the gap of diversity control while maintaining consistency. 

\section{Conclusion}
We presented Adaptive Prompt Pruning (APP), a simple yet effective method for controlling diversity in LLM-agent simulations. By leveraging attention scores to identify and prune overemphasized prompt segments, APP enables fine-grained diversity control via a single parameter and complements existing decoding-based techniques. Our analysis reveals that prompt components, especially high-attention segments, significantly constrain generation. APP offers a flexible framework for balancing consistency and creativity in multi-agent settings, paving the way for more robust and controllable simulations.

\section*{Limitations}
\label{sec:limitations}

First, the obtained results pertain to the two datasets used in this study. Although we believe that \texttt{GA} and \texttt{HA} cover a general context for agent simulation, there may be cases beyond the scope of our discussion.

Second, an additional compute budget and the availability of attention weights are required during inference. To further optimize time efficiency, focusing pruning efforts
solely on critical utterances instead of processing every utterance may be a
promising future direction.

Finally, despite promising results on revision, several directions warrant further exploration. First, investigating potential biases in the LLM's judgments and their correlation with dialogue diversity presents a valuable avenue for future research. Second, attention is needed for utterances that are difficult to revise solely by rolling back, such as when the agent is asked, ``What is your major?" and lacks relevant information to respond faithfully. Drawing on the distinction between discrimination and criticism \cite{saunders2022self}, the LLM could be queried to assess its ability to ``know'' the appropriate revision direction using the removed units. If capable, a comment-based modification could be applied; otherwise, rolling back could be used to benefit from diversity in generation. Combining these two approaches may improve pipeline efficiency.

\section*{Acknowledgments}
This work was supported by JST-CRONOS Grant Number JPMJCS24K8, JSPS KAKENHI Grant Number JP23K28139, and the Institute of AI and Beyond of the University of Tokyo.

\bibliography{3_custom}

\begin{thebibliography}{40}
\providecommand{\natexlab}[1]{#1}

\bibitem[{Ackley et~al.(1985)Ackley, Hinton, and Sejnowski}]{ackley1985learning}
David~H Ackley, Geoffrey~E Hinton, and Terrence~J Sejnowski. 1985.
\newblock A learning algorithm for boltzmann machines.
\newblock \emph{Cognitive science}, 9(1):147--169.

\bibitem[{Chen et~al.(2024{\natexlab{a}})Chen, Su, Zuo, Yang, Yuan, Chan, Yu, Lu, Hung, Qian, Qin, Cong, Xie, Liu, Sun, and Zhou}]{chen2024agentverse}
Weize Chen, Yusheng Su, Jingwei Zuo, Cheng Yang, Chenfei Yuan, Chi-Min Chan, Heyang Yu, Yaxi Lu, Yi-Hsin Hung, Chen Qian, Yujia Qin, Xin Cong, Ruobing Xie, Zhiyuan Liu, Maosong Sun, and Jie Zhou. 2024{\natexlab{a}}.
\newblock \href {https://openreview.net/forum?id=EHg5GDnyq1} {Agentverse: Facilitating multi-agent collaboration and exploring emergent behaviors}.
\newblock In \emph{The Twelfth International Conference on Learning Representations}.

\bibitem[{Chen et~al.(2024{\natexlab{b}})Chen, Chi, Wang, and Zhou}]{chen2024premise}
Xinyun Chen, Ryan~Andrew Chi, Xuezhi Wang, and Denny Zhou. 2024{\natexlab{b}}.
\newblock Premise order matters in reasoning with large language models.
\newblock In \emph{Forty-first International Conference on Machine Learning}.

\bibitem[{Cheng et~al.(2024)Cheng, Zhang, Zhang, Meng, Hong, Li, Wang, Wang, Yin, Zhao et~al.}]{agent_survey_exp_24}
Yuheng Cheng, Ceyao Zhang, Zhengwen Zhang, Xiangrui Meng, Sirui Hong, Wenhao Li, Zihao Wang, Zekai Wang, Feng Yin, Junhua Zhao, et~al. 2024.
\newblock Exploring large language model based intelligent agents: Definitions, methods, and prospects.
\newblock \emph{arXiv preprint arXiv:2401.03428}.

\bibitem[{Chu et~al.(2024)Chu, Chen, and Nakayama}]{chu2024cohesive}
KuanChao Chu, Yi-Pei Chen, and Hideki Nakayama. 2024.
\newblock \href {https://openreview.net/forum?id=3ypWPhMGhV} {Cohesive conversations: Enhancing authenticity in multi-agent simulated dialogues}.
\newblock In \emph{First Conference on Language Modeling}.

\bibitem[{Chung et~al.(2023)Chung, Kamar, and Amershi}]{chung-etal-2023-increasing}
John Chung, Ece Kamar, and Saleema Amershi. 2023.
\newblock \href {https://doi.org/10.18653/v1/2023.acl-long.34} {Increasing diversity while maintaining accuracy: Text data generation with large language models and human interventions}.
\newblock In \emph{Proceedings of the 61st Annual Meeting of the Association for Computational Linguistics (Volume 1: Long Papers)}, pages 575--593. Association for Computational Linguistics.

\bibitem[{Dodge et~al.(2021)Dodge, Sap, Marasovi{\'c}, Agnew, Ilharco, Groeneveld, Mitchell, and Gardner}]{c4dataset_21}
Jesse Dodge, Maarten Sap, Ana Marasovi{\'c}, William Agnew, Gabriel Ilharco, Dirk Groeneveld, Margaret Mitchell, and Matt Gardner. 2021.
\newblock Documenting large webtext corpora: A case study on the colossal clean crawled corpus.
\newblock In \emph{Proceedings of the 2021 Conference on Empirical Methods in Natural Language Processing}, pages 1286--1305. Association for Computational Linguistics.

\bibitem[{Dubey et~al.(2024)Dubey, Jauhri, Pandey, Kadian, Al-Dahle, Letman, Mathur, Schelten, Yang, Fan et~al.}]{llama3_24}
Abhimanyu Dubey, Abhinav Jauhri, Abhinav Pandey, Abhishek Kadian, Ahmad Al-Dahle, Aiesha Letman, Akhil Mathur, Alan Schelten, Amy Yang, Angela Fan, et~al. 2024.
\newblock The llama 3 herd of models.
\newblock \emph{arXiv preprint arXiv:2407.21783}.

\bibitem[{Gao et~al.(2024)Gao, Xu, Chen, Wang, He, and Li}]{gao2024simulating}
Chen Gao, Fengli Xu, Xu~Chen, Xiang Wang, Xiangnan He, and Yong Li. 2024.
\newblock \href {https://doi.org/10.1145/3589335.3641253} {Simulating human society with large language model agents: City, social media, and economic system}.
\newblock In \emph{Companion Proceedings of the ACM Web Conference 2024}, WWW '24, page 1290–1293, New York, NY, USA. Association for Computing Machinery.

\bibitem[{{Gemma Team}(2024)}]{gemma_2024}
{Gemma Team}. 2024.
\newblock \href {https://doi.org/10.34740/KAGGLE/M/3301} {Gemma}.

\bibitem[{Guo et~al.(2024)Guo, Chen, Wang, Chang, Pei, Chawla, Wiest, and Zhang}]{ijcai2024p890}
Taicheng Guo, Xiuying Chen, Yaqi Wang, Ruidi Chang, Shichao Pei, Nitesh~V. Chawla, Olaf Wiest, and Xiangliang Zhang. 2024.
\newblock \href {https://doi.org/10.24963/ijcai.2024/890} {Large language model based multi-agents: A survey of progress and challenges}.
\newblock In \emph{Proceedings of the Thirty-Third International Joint Conference on Artificial Intelligence, {IJCAI-24}}, pages 8048--8057. International Joint Conferences on Artificial Intelligence Organization.
\newblock Survey Track.

\bibitem[{Holtzman et~al.(2020)Holtzman, Buys, Du, Forbes, and Choi}]{Holtzman2020The}
Ari Holtzman, Jan Buys, Li~Du, Maxwell Forbes, and Yejin Choi. 2020.
\newblock \href {https://openreview.net/forum?id=rygGQyrFvH} {The curious case of neural text degeneration}.
\newblock In \emph{International Conference on Learning Representations}.

\bibitem[{Hong et~al.(2024)Hong, Zhuge, Chen, Zheng, Cheng, Wang, Zhang, Wang, Yau, Lin, Zhou, Ran, Xiao, Wu, and Schmidhuber}]{hong2024metagpt}
Sirui Hong, Mingchen Zhuge, Jonathan Chen, Xiawu Zheng, Yuheng Cheng, Jinlin Wang, Ceyao Zhang, Zili Wang, Steven Ka~Shing Yau, Zijuan Lin, Liyang Zhou, Chenyu Ran, Lingfeng Xiao, Chenglin Wu, and J{\"u}rgen Schmidhuber. 2024.
\newblock \href {https://openreview.net/forum?id=VtmBAGCN7o} {Meta{GPT}: Meta programming for a multi-agent collaborative framework}.
\newblock In \emph{The Twelfth International Conference on Learning Representations}.

\bibitem[{Jiang et~al.(2023)Jiang, Sablayrolles, Mensch, Bamford, Chaplot, de~las Casas, Bressand, Lengyel, Lample, Saulnier, Lavaud, Lachaux, Stock, Scao, Lavril, Wang, Lacroix, and Sayed}]{mistral7b}
Albert~Q. Jiang, Alexandre Sablayrolles, Arthur Mensch, Chris Bamford, Devendra~Singh Chaplot, Diego de~las Casas, Florian Bressand, Gianna Lengyel, Guillaume Lample, Lucile Saulnier, Lélio~Renard Lavaud, Marie-Anne Lachaux, Pierre Stock, Teven~Le Scao, Thibaut Lavril, Thomas Wang, Timothée Lacroix, and William~El Sayed. 2023.
\newblock \href {https://arxiv.org/abs/2310.06825} {Mistral 7b}.
\newblock \emph{Preprint}, arXiv:2310.06825.

\bibitem[{Lee et~al.(2022)Lee, Ping, Xu, Patwary, Fung, Shoeybi, and Catanzaro}]{lee2022factuality}
Nayeon Lee, Wei Ping, Peng Xu, Mostofa Patwary, Pascale~N Fung, Mohammad Shoeybi, and Bryan Catanzaro. 2022.
\newblock Factuality enhanced language models for open-ended text generation.
\newblock \emph{Advances in Neural Information Processing Systems}, 35:34586--34599.

\bibitem[{Li et~al.(2016)Li, Galley, Brockett, Gao, and Dolan}]{li-etal-2016-diversity}
Jiwei Li, Michel Galley, Chris Brockett, Jianfeng Gao, and Bill Dolan. 2016.
\newblock \href {https://doi.org/10.18653/v1/N16-1014} {A diversity-promoting objective function for neural conversation models}.
\newblock In \emph{Proceedings of the 2016 Conference of the North {A}merican Chapter of the Association for Computational Linguistics: Human Language Technologies}, pages 110--119, San Diego, California. Association for Computational Linguistics.

\bibitem[{Liu et~al.(2024{\natexlab{a}})Liu, Lin, Hewitt, Paranjape, Bevilacqua, Petroni, and Liang}]{liu2024lost}
Nelson~F Liu, Kevin Lin, John Hewitt, Ashwin Paranjape, Michele Bevilacqua, Fabio Petroni, and Percy Liang. 2024{\natexlab{a}}.
\newblock Lost in the middle: How language models use long contexts.
\newblock \emph{Transactions of the Association for Computational Linguistics}, 12:157--173.

\bibitem[{Liu et~al.(2024{\natexlab{b}})Liu, Zhang, Li, Liu, and Yang}]{liu2024a}
Zijun Liu, Yanzhe Zhang, Peng Li, Yang Liu, and Diyi Yang. 2024{\natexlab{b}}.
\newblock \href {https://openreview.net/forum?id=XII0Wp1XA9} {A dynamic {LLM}-powered agent network for task-oriented agent collaboration}.
\newblock In \emph{First Conference on Language Modeling}.

\bibitem[{McCoy et~al.(2023)McCoy, Yao, Friedman, Hardy, and Griffiths}]{ember_23}
R~Thomas McCoy, Shunyu Yao, Dan Friedman, Matthew Hardy, and Thomas~L Griffiths. 2023.
\newblock Embers of autoregression: Understanding large language models through the problem they are trained to solve.
\newblock \emph{arXiv preprint arXiv:2309.13638}.

\bibitem[{Mendon{\c{c}}a et~al.(2024)Mendon{\c{c}}a, Lavie, and Trancoso}]{mendonca-etal-2024-benchmarking}
John Mendon{\c{c}}a, Alon Lavie, and Isabel Trancoso. 2024.
\newblock \href {https://aclanthology.org/2024.nlp4convai-1.1/} {On the benchmarking of {LLM}s for open-domain dialogue evaluation}.
\newblock In \emph{Proceedings of the 6th Workshop on NLP for Conversational AI (NLP4ConvAI 2024)}, pages 1--12, Bangkok, Thailand. Association for Computational Linguistics.

\bibitem[{{Meta AI}(2024)}]{llama31_huggingface}
{Meta AI}. 2024.
\newblock \href {https://huggingface.co/meta-llama/Llama-3.1-8B} {Llama 3.1 - 8b}.
\newblock Available at: \url{https://huggingface.co/meta-llama/Llama-3.1-8B}.

\bibitem[{Pan et~al.(2023)Pan, Saxon, Xu, Nathani, Wang, and Wang}]{pan2023automaticallycorrectinglargelanguage}
Liangming Pan, Michael Saxon, Wenda Xu, Deepak Nathani, Xinyi Wang, and William~Yang Wang. 2023.
\newblock \href {https://arxiv.org/abs/2308.03188} {Automatically correcting large language models: Surveying the landscape of diverse self-correction strategies}.
\newblock \emph{Preprint}, arXiv:2308.03188.

\bibitem[{Panickssery et~al.(2024)Panickssery, Bowman, and Feng}]{panickssery2024llm}
Arjun Panickssery, Samuel~R. Bowman, and Shi Feng. 2024.
\newblock \href {https://openreview.net/forum?id=4NJBV6Wp0h} {{LLM} evaluators recognize and favor their own generations}.
\newblock In \emph{The Thirty-eighth Annual Conference on Neural Information Processing Systems}.

\bibitem[{Park et~al.(2023)Park, O'Brien, Cai, Morris, Liang, and Bernstein}]{generative_agent23}
Joon~Sung Park, Joseph O'Brien, Carrie~Jun Cai, Meredith~Ringel Morris, Percy Liang, and Michael~S. Bernstein. 2023.
\newblock \href {https://doi.org/10.1145/3586183.3606763} {Generative agents: Interactive simulacra of human behavior}.
\newblock In \emph{Proceedings of the 36th Annual ACM Symposium on User Interface Software and Technology}, UIST '23.

\bibitem[{Park et~al.(2024)Park, Zou, Shaw, Hill, Cai, Morris, Willer, Liang, and Bernstein}]{park24_ganext}
Joon~Sung Park, Carolyn~Q Zou, Aaron Shaw, Benjamin~Mako Hill, Carrie Cai, Meredith~Ringel Morris, Robb Willer, Percy Liang, and Michael~S Bernstein. 2024.
\newblock Generative agent simulations of 1,000 people.
\newblock \emph{arXiv preprint arXiv:2411.10109}.

\bibitem[{Reimers(2019)}]{reimers2019sentence}
N~Reimers. 2019.
\newblock Sentence-bert: Sentence embeddings using siamese bert-networks.
\newblock \emph{arXiv preprint arXiv:1908.10084}.

\bibitem[{{Replica Inc.}(2023)}]{replica}
{Replica Inc.} 2023.
\newblock \href {https://www.replicastudios.com/blog/smart-npc-plugin-release} {Replica smart npcs}.
\newblock Available at: \url{https://www.replicastudios.com/blog/smart-npc-plugin-release}.

\bibitem[{Saunders et~al.(2022)Saunders, Yeh, Wu, Bills, Ouyang, Ward, and Leike}]{saunders2022self}
William Saunders, Catherine Yeh, Jeff Wu, Steven Bills, Long Ouyang, Jonathan Ward, and Jan Leike. 2022.
\newblock Self-critiquing models for assisting human evaluators.
\newblock \emph{arXiv preprint arXiv:2206.05802}.

\bibitem[{Stasaski et~al.(2020)Stasaski, Yang, and Hearst}]{stasaski-etal-2020-diverse}
Katherine Stasaski, Grace~Hui Yang, and Marti~A. Hearst. 2020.
\newblock \href {https://doi.org/10.18653/v1/2020.acl-main.446} {More diverse dialogue datasets via diversity-informed data collection}.
\newblock In \emph{Proceedings of the 58th Annual Meeting of the Association for Computational Linguistics}, pages 4958--4968, Online. Association for Computational Linguistics.

\bibitem[{Sumers et~al.(2024)Sumers, Yao, Narasimhan, and Griffiths}]{sumers2024cognitive}
Theodore Sumers, Shunyu Yao, Karthik Narasimhan, and Thomas Griffiths. 2024.
\newblock \href {https://openreview.net/forum?id=1i6ZCvflQJ} {Cognitive architectures for language agents}.
\newblock \emph{Transactions on Machine Learning Research}.

\bibitem[{Thakur et~al.(2024)Thakur, Choudhary, Ramayapally, Vaidyanathan, and Hupkes}]{thakur2024judging}
Aman~Singh Thakur, Kartik Choudhary, Venkat~Srinik Ramayapally, Sankaran Vaidyanathan, and Dieuwke Hupkes. 2024.
\newblock Judging the judges: Evaluating alignment and vulnerabilities in llms-as-judges.
\newblock \emph{arXiv preprint arXiv:2406.12624}.

\bibitem[{Wang et~al.(2021)Wang, Bao, Huang, Dong, and Wei}]{wang2021minilmv2}
Wenhui Wang, Hangbo Bao, Shaohan Huang, Li~Dong, and Furu Wei. 2021.
\newblock Minilmv2: Multi-head self-attention relation distillation for compressing pretrained transformers.
\newblock In \emph{Findings of the Association for Computational Linguistics: ACL-IJCNLP 2021}, pages 2140--2151.

\bibitem[{Wang et~al.(2023)Wang, Chiu, and Chiu}]{humanoid_23}
Zhilin Wang, Yu~Ying Chiu, and Yu~Cheung Chiu. 2023.
\newblock Humanoid agents: Platform for simulating human-like generative agents.
\newblock In \emph{Proceedings of the 2023 Conference on Empirical Methods in Natural Language Processing: System Demonstrations}, pages 167--176.

\bibitem[{Weston and Sukhbaatar(2023)}]{system2_23}
Jason Weston and Sainbayar Sukhbaatar. 2023.
\newblock System 2 attention (is something you might need too).
\newblock \emph{arXiv preprint arXiv:2311.11829}.

\bibitem[{Xi et~al.(2023)Xi, Chen, Guo, He, Ding, Hong, Zhang, Wang, Jin, Zhou et~al.}]{agent_survey_rise_23}
Zhiheng Xi, Wenxiang Chen, Xin Guo, Wei He, Yiwen Ding, Boyang Hong, Ming Zhang, Junzhe Wang, Senjie Jin, Enyu Zhou, et~al. 2023.
\newblock The rise and potential of large language model based agents: A survey.
\newblock \emph{arXiv preprint arXiv:2309.07864}.

\bibitem[{Yang et~al.(2025)Yang, Yang, Zhang, Hui, Zheng, Yu, Li, Liu, Huang, Wei, Lin, Yang, Tu, Zhang, Yang, Yang, Zhou, Lin, Dang, Lu, Bao, Yang, Yu, Li, Xue, Zhang, Zhu, Men, Lin, Li, Tang, Xia, Ren, Ren, Fan, Su, Zhang, Wan, Liu, Cui, Zhang, and Qiu}]{qwen25technicalreport}
An~Yang, Baosong Yang, Beichen Zhang, Binyuan Hui, Bo~Zheng, Bowen Yu, Chengyuan Li, Dayiheng Liu, Fei Huang, Haoran Wei, Huan Lin, Jian Yang, Jianhong Tu, Jianwei Zhang, Jianxin Yang, Jiaxi Yang, Jingren Zhou, Junyang Lin, Kai Dang, Keming Lu, Keqin Bao, Kexin Yang, Le~Yu, Mei Li, Mingfeng Xue, Pei Zhang, Qin Zhu, Rui Men, Runji Lin, Tianhao Li, Tianyi Tang, Tingyu Xia, Xingzhang Ren, Xuancheng Ren, Yang Fan, Yang Su, Yichang Zhang, Yu~Wan, Yuqiong Liu, Zeyu Cui, Zhenru Zhang, and Zihan Qiu. 2025.
\newblock \href {https://arxiv.org/abs/2412.15115} {Qwen2.5 technical report}.
\newblock \emph{Preprint}, arXiv:2412.15115.

\bibitem[{Yao et~al.(2023)Yao, Yu, Zhao, Shafran, Griffiths, Cao, and Narasimhan}]{treeofthought_23}
Shunyu Yao, Dian Yu, Jeffrey Zhao, Izhak Shafran, Tom Griffiths, Yuan Cao, and Karthik Narasimhan. 2023.
\newblock Tree of thoughts: Deliberate problem solving with large language models.
\newblock \emph{Advances in Neural Information Processing Systems}.

\bibitem[{Zheng et~al.(2023)Zheng, Chiang, Sheng, Zhuang, Wu, Zhuang, Lin, Li, Li, Xing, Zhang, Gonzalez, and Stoica}]{zheng2023judging}
Lianmin Zheng, Wei-Lin Chiang, Ying Sheng, Siyuan Zhuang, Zhanghao Wu, Yonghao Zhuang, Zi~Lin, Zhuohan Li, Dacheng Li, Eric Xing, Hao Zhang, Joseph~E. Gonzalez, and Ion Stoica. 2023.
\newblock \href {https://openreview.net/forum?id=uccHPGDlao} {Judging {LLM}-as-a-judge with {MT}-bench and chatbot arena}.
\newblock In \emph{Thirty-seventh Conference on Neural Information Processing Systems Datasets and Benchmarks Track}.

\bibitem[{Zhou et~al.(2023)Zhou, Pang, Shen, and Cheng}]{zhou-etal-2023-simoap}
Junkai Zhou, Liang Pang, Huawei Shen, and Xueqi Cheng. 2023.
\newblock \href {https://doi.org/10.18653/v1/2023.acl-long.553} {{S}im{OAP}: Improve coherence and consistency in persona-based dialogue generation via over-sampling and post-evaluation}.
\newblock In \emph{Proceedings of the 61st Annual Meeting of the Association for Computational Linguistics (Volume 1: Long Papers)}, pages 9945--9959, Toronto, Canada. Association for Computational Linguistics.

\bibitem[{Zhou et~al.(2024)Zhou, Zhu, Mathur, Zhang, Yu, Qi, Morency, Bisk, Fried, Neubig, and Sap}]{zhou2024sotopia}
Xuhui Zhou, Hao Zhu, Leena Mathur, Ruohong Zhang, Haofei Yu, Zhengyang Qi, Louis-Philippe Morency, Yonatan Bisk, Daniel Fried, Graham Neubig, and Maarten Sap. 2024.
\newblock \href {https://openreview.net/forum?id=mM7VurbA4r} {{SOTOPIA}: Interactive evaluation for social intelligence in language agents}.
\newblock In \emph{The Twelfth International Conference on Learning Representations}.

\end{thebibliography}

\clearpage
\appendix
\appendix

\section{Details of the Datasets}
\subsection{Dialogue Cases}
Table~\ref{tab:caselist} lists the 20 cases used in this study.

\begin{table}[!thp]
\centering
\small
\resizebox{1.0\linewidth}{!}{
\begin{tabular}{lll}
\toprule

Time Stamp & Agent A & Agent B  \\ 

\midrule
2023-02-13 07:40:50 & Tamara Taylor & Carmen Ortiz \\
2023-02-13 09:00:40 & Arthur Burton & Sam Moore \\
2023-02-13 09:46:20 & Francisco Lopez & Abigail Chen \\
2023-02-13 10:21:20 & John Lin & Tom Moreno \\
2023-02-13 11:03:40 & Giorgio Rossi & Klaus Mueller \\
2023-02-13 11:10:40 & Arthur Burton & Ryan Park \\
2023-02-13 12:23:50 & Hailey Johnson & Giorgio Rossi \\
2023-02-13 12:28:10 & Sam Moore & Yuriko Yamamoto \\
2023-02-13 13:09:10 & Ayesha Khan & Mei Lin \\
2023-02-13 13:33:20 & Sam Moore & Abigail Chen \\
2023-02-13 14:28:10 & Carmen Ortiz & Rajiv Patel \\
2023-02-13 14:46:50 & Maria Lopez & Ayesha Khan \\
2023-02-13 15:05:20 & Jennifer Moore & Tamara Taylor \\
2023-02-13 15:36:50 & Ayesha Khan & Wolfgang Schulz \\
2023-02-13 15:53:50 & Ayesha Khan & Mei Lin \\
2023-02-13 16:44:20 & Carmen Ortiz & Latoya Williams \\
2023-02-13 17:18:20 & Maria Lopez & Ayesha Khan \\
2023-02-13 17:27:00 & Mei Lin & Eddy Lin \\
2023-02-13 19:36:20 & Francisco Lopez & Rajiv Patel \\
2023-02-13 20:04:40 & Rajiv Patel & Hailey Johnson \\

\bottomrule
\end{tabular}
}
\caption{Cases sampled from \citet{generative_agent23} for our study.}
\label{tab:caselist}
\end{table}

\subsection{Data Access for \texttt{GA}}
The simulation logs of \cite{generative_agent23} can be accessed from the following URL.

{\small \verb|https://reverie.herokuapp.com/arXiv_Demo/|
}

\subsection{Implementation Details for \texttt{HA}}
As described in the main paper, \texttt{HA} extends \texttt{GA} by introducing a human needs block. This block captures three types of information: basic needs, emotions, and relationship closeness.

\begin{itemize}
    \item \textbf{Basic needs}: These include five states—fullness, social, fun, health, and energy—each corresponding to an unsatisfied adjective: hungry, lonely, bored, unwell, and tired. In the original paper, these states are represented by values ranging from 0 to 10. A state is considered unsatisfied when its value falls below 4. When this occurs, the following item is added to the block: ``Agent A is \{modifier\} \{unsatisfied adjective\}.'' Modifier includes: ``slightly'', ``'', ``very'', and ``extremely''. 

    \item \textbf{Emotions}: Emotional states include disgusted, afraid, sad, surprised, happy, angry, and neutral. If the emotional state is not neutral, the following item is added to the block: ``Agent A is feeling extremely \{emotion\}.''
    
    \item \textbf{Relationship closeness}: Based on the relationship between speaker and listener, the following item is added to the block: ``Agent A is feeling \{closeness level\} to Agent B.'' Closeness levels are distant, rather close, close, and very close.
\end{itemize}

In our implementation, agent states are sampled probabilistically:
\begin{itemize}
    \item Basic needs: 40\% chance of being unsatisfied (20\% for energy), with modifiers assigned equally.
    \item Emotions: Each non-``neutral'' emotion has an 8\% chance of selection.
    \item Relationship closeness: The probabilities are distributed as 50\%, 20\%, 20\%, and 10\%, respectively.
\end{itemize}

For each case, the human needs of each agent are sampled independently (using separate seeds for agents), and remain constant within a single case.

\section{Prompt and Samples}
Table~\ref{tab:prompt} presents an example prompt and their composition used for utterance generation in our study. The wording of the content has been modified or adopted from \cite{generative_agent23,humanoid_23}. Table~\ref{tab:rmatt85_sample} and Table~\ref{tab:rmatt0_sample} shows sample dialogues under a same initial condition for lambda=0.85 and 0.0.

\begin{figure}[!th]
  \centering
  \begin{subfigure}[t]{0.75\linewidth}
      \includegraphics[width=\linewidth]{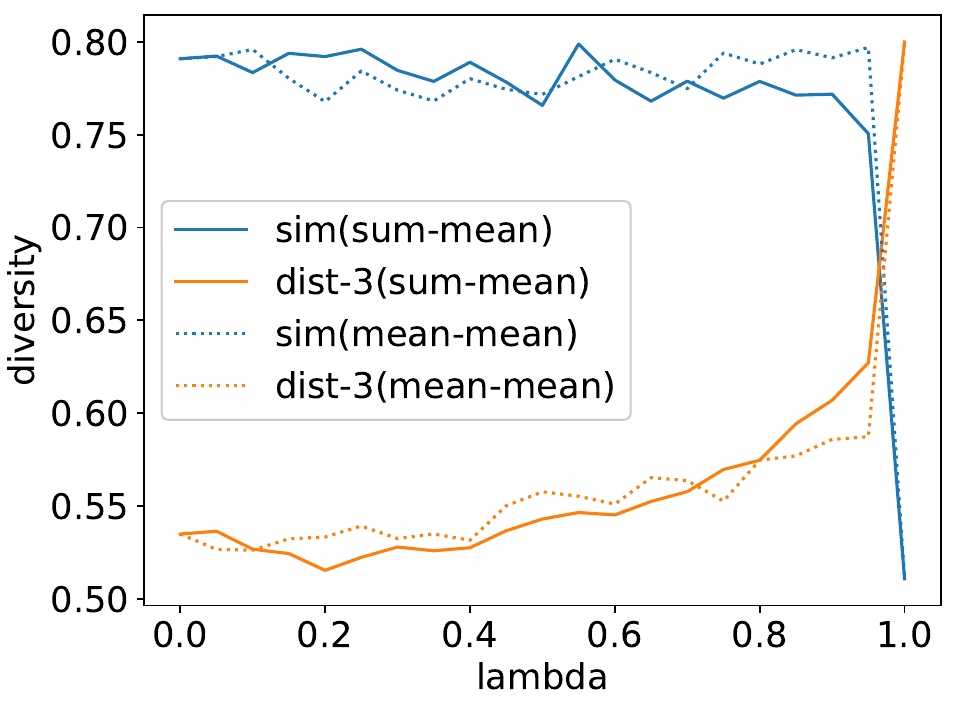}
  \end{subfigure}
  \caption{Lambda vs. diversity for different reducer choices.}
  \label{fig:rmatt_analysis_appendix}
\end{figure}

\section{Additional Study on Adaptive Prompt Pruning}
\label{appendix:reducer}
\paragraph{Reducer}
We evaluate the effects of different reducers $R(\cdot)$ on the selection results of unit removal. As shown in Fig.~\ref{fig:rmatt_analysis_appendix}, the solid line represents the ``sum-mean'' method we adopted, while the dotted lines correspond to the ``mean-mean'' method. Unlike ``sum-mean'', the ``mean-mean'' method averages the attention scores across all tokens within a unit instead of summing them, thereby reducing the scoring advantage of longer units. However, we observe that ``mean-mean'' achieves inferior improvements in diversity when the $\lambda$ value is large. Additionally, changes under the sim metric initially decrease and then increase, indicating a weaker linear relationship with $\lambda$. Given the goal of serving as a control parameter, we argue that the ``sum-mean'' method, which preserves the length bias of units, is a more suitable choice.

\section{Extended Analysis on Removal}
\label{appendix:analysis_removal}
We present additional perspectives to deepen the understanding of the unit removal method. Since the results for ``Remove memory (RMm)'' exhibit the most significant differences, we use this setting as a representative case to conduct the following experiments.

\begin{figure}[!th]
  \centering
    \begin{subfigure}{0.49\linewidth}  
        \includegraphics[width=\linewidth]{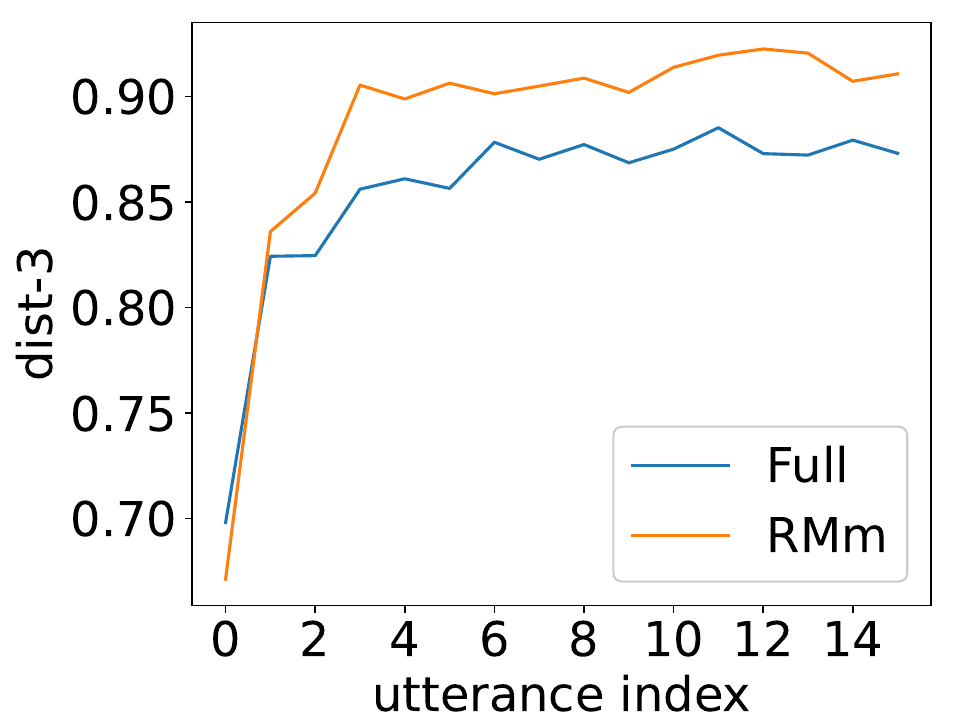}
        \caption{utterance index vs. dist-3}
        \label{fig:perutt_ng}
    \end{subfigure}
    \begin{subfigure}{0.49\linewidth}
        \includegraphics[width=\linewidth]{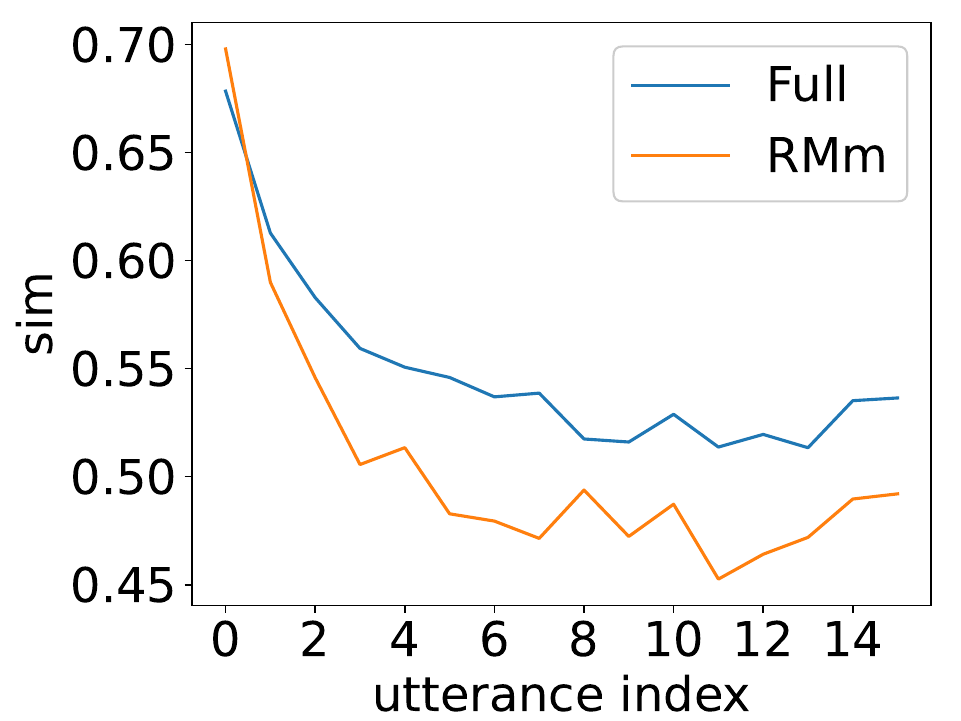}
        \caption{utterance index vs. sim}
        \label{fig:perutt_sim}
    \end{subfigure}
  \caption{Tracking the progression of dialogue diversity through per-utterance measures.}
  \label{fig:perutt}
\end{figure}

\paragraph{Diversity improvement is driven by the first few rounds}
In this paper, we examined diversity across different dialogue trials. But at what point does the divergence between dialogues occur? Using utterances as the unit of analysis, we calculated the diversity of utterances at corresponding positions across dialogues, employing the same similarity and Dist-N metrics. As shown in Fig.~\ref{fig:perutt}, we compared the differences between the full prompt and RMm. Regardless of whether the removal operation was applied, diversity consistently increased during the initial rounds of dialogue, with index 1 (the listener’s first response) being particularly critical. Building on this foundation, RMm further amplifies its divergence from the full prompt around indices 2 to 3, before stabilizing in the later stages of the dialogue.

\paragraph{Measuring the exclusiveness of content across settings} 
After applying the RMm setting, the diversity among different trials increases significantly. To further investigate whether RMm generates more novel content or whether most of the generated content overlaps with the dialogues produced under the full-prompt setting, we measure the exclusiveness of the generated dialogues between two settings. 
Given N dialogues generated under settings A and B, respectively, we compute the following metrics:
\begin{enumerate}
    \item Average B-to-A max similarity (Avg max sim): The average of the maximum similarity scores for each dialogue in B compared to the dialogues in A. 
    \item Exclusive unique n-gram ratio for B (Excl-n): The proportion of unique n-grams in all dialogues of B that do not appear in A.
\end{enumerate}
The calculations for similarity and unique n-grams follow the same methodology used in this study. We compare the differences between the full-to-full (averaged over three different seeds) and full-to-RMm settings, with the results presented in Table~\ref{tab:exclusive}. These findings indicate that RMm indeed generates more exclusive content.

\begin{table}[!thp]
\centering
\small
\resizebox{1.0\linewidth}{!}{
\begin{tabular}{lcccc}
\toprule

& Avg max sim ($\downarrow$) & Excl-1 & Excl-2 & Excl-3 ($\uparrow$) \\ 

\midrule
Full to Full & 0.881 & 0.382 & 0.580 & 0.720 \\
Full to RMm & 0.815 & 0.484 & 0.699 & 0.814 \\

\bottomrule
\end{tabular}
}
\caption{Exclusiveness measure: RMm performs better in these metrics, demonstrating its ability to generate novel content compared to Full.}
\label{tab:exclusive}
\end{table}

\begin{table*}[htbp]
\centering
\small
\begin{tabular}{l c cccc ccc}
\toprule
\multirow{2}{*}{Model} & \multirow{2}{*}{$\lambda$} & \multicolumn{4}{c}{Diversity Metrics} & \multicolumn{3}{c}{Other Stats} \\
\cmidrule(lr){3-6} \cmidrule(lr){7-9}
 & & Sim ($\downarrow$) & Dist-1 & Dist-2 & Dist-3 ($\uparrow$) & Turns & Utt. length & Last turn rep. rate \\
\midrule
Qwen & 0.0 & 0.762 & 0.120 & 0.456 & 0.668 & 13.7 & 43.4 & 0.11 \\
Qwen & 0.4 & 0.755 & 0.119 & 0.444 & 0.651 & 13.1 & 46.8 & 0.16 \\
Qwen & 0.6 & 0.724 & 0.124 & 0.446 & 0.648 & 11.9 & 46.5 & 0.19 \\
Qwen & 0.7 & 0.739 & 0.118 & 0.430 & 0.628 & 12.6 & 45.6 & 0.24 \\
Qwen & 0.9 & 0.733 & 0.115 & 0.413 & 0.599 & 13.1 & 44.1 & 0.39 \\
Qwen & 1.0 & \textbf{0.597} & \textbf{0.161} & \textbf{0.529} & \textbf{0.714} & 14.5 & 32.4 & 0.33 \\
\midrule
gemma & 0.0 & 0.712 & 0.190 & 0.589 & 0.778 & 13.9 & 26.5 & 0.02 \\
gemma & 0.3 & 0.687 & 0.191 & 0.594 & 0.789 & 14.5 & 27.4 & 0.03 \\
gemma & 0.5 & 0.668 & 0.205 & 0.618 & \textbf{0.807} & 13.7 & 25.6 & 0.04 \\
gemma & 0.7 & 0.678 & 0.195 & 0.601 & 0.794 & 14.0 & 26.4 & 0.08 \\
gemma & 0.9 & 0.699 & 0.194 & 0.596 & 0.788 & 13.6 & 26.0 & 0.06 \\
gemma & 1.0 & \textbf{0.629} & \textbf{0.226} & \textbf{0.621} & 0.790 & 14.9 & 17.5 & 0.02 \\
\midrule
Mistral & 0.0 & 0.691 & 0.219 & 0.560 & 0.724 & 4.2 & 43.0 & 0.00 \\
Mistral & 0.3 & 0.677 & 0.218 & 0.570 & 0.741 & 4.2 & 45.9 & 0.00 \\
Mistral & 0.5 & 0.688 & 0.192 & 0.534 & 0.714 & 4.9 & 48.9 & 0.00 \\
Mistral & 0.7 & 0.681 & 0.186 & 0.529 & 0.712 & 4.9 & 49.6 & 0.00 \\
Mistral & 0.9 & 0.698 & 0.183 & 0.529 & 0.715 & 4.9 & 49.3 & 0.00 \\
Mistral & 1.0 & \textbf{0.668} & \textbf{0.252} & \textbf{0.595} & \textbf{0.745} & 5.0 & 29.1 & 0.00 \\

\bottomrule
\end{tabular}
\caption{Results of applying our method (APP) to various LLMs at critical lambda values. Diversity metrics are measured as in the main paper. Additional statistics include the average number of dialogue turns, average utterance length, and the repetition rate of the final utterance in each dialogue.}
\label{tab:nonllama}
\end{table*}

\section{Additional Study on Other Factors Affecting Diversity}
\label{appendix:analysis_other_factors}
\paragraph{Effect of block length} 
We simulate variations in block length by randomly duplicating or deleting items within the blocks. The word count for each block containing items is adjusted to either 250 or 750 words (\texttt{BLN250} and \texttt{BLN750}). For blocks other than memory, these operations effectively result in either an increase or no change in length. To isolate the effect of memory, we exclude it from the analysis. The results indicate that, compared to RMm, \texttt{BLN250}+RMm exhibits minimal differences in diversity, whereas \texttt{BLN750}+RMm shows a significant decline in the Dist-N metric. This finding underscores the detrimental impact of excessive redundant content on diversity.

\begin{table}[!thp]
\centering
\small
\resizebox{1.0\linewidth}{!}{
\begin{tabular}{lcccc}
\toprule

& Sim ($\downarrow$) & Dist-1 & Dist-2 & Dist-3 ($\uparrow$) \\ 
\midrule
Full       & 0.791 & 0.095 & 0.350 & 0.535 \\
\midrule
& & Length & & \\
\midrule
RMm        & 0.736 & 0.119 & 0.429 & 0.636 \\
BLN250+RMm & 0.734 & 0.118 & 0.423 & 0.627 \\
BLN750+RMm & 0.744 & 0.110 & 0.377 & 0.556 \\

\bottomrule
\end{tabular}
}
\caption{Diversity changes resulting from altering block length in the text space.}
\label{tab:explore_appendix}
\end{table}

\section{Other Backbone LLMs}
\label{sec:appendix_other_llm}
In addition to the LLaMA 3 and LLaMA 3.1 models presented in the main text, we further evaluated our method on three additional backbone models. 
The results are summarized in Table~\ref{tab:nonllama}. 
Overall, APP demonstrates solid performance across these models, effectively controlling and enhancing dialogue diversity.

\paragraph{Models}
We selected three popular open-source model series: Qwen2.5-7B-Instruct \cite{qwen25technicalreport}, gemma-2-9b-it \cite{gemma_2024}, and Mistral-7B-Instruct-v0.3 \cite{mistral7b}. These models were loaded in FP32 (Qwen) or BF16, and the decoding settings (i.e., temperature, top-p, and top-k) were kept consistent with those used in the main paper. In our implementation, we observed that one token per item might be omitted during attention score computation for Qwen. However, this had a negligible impact on the overall results.

\begin{figure}[!th]
  \centering
  \begin{subfigure}[t]{0.75\linewidth}
      \includegraphics[width=\linewidth]{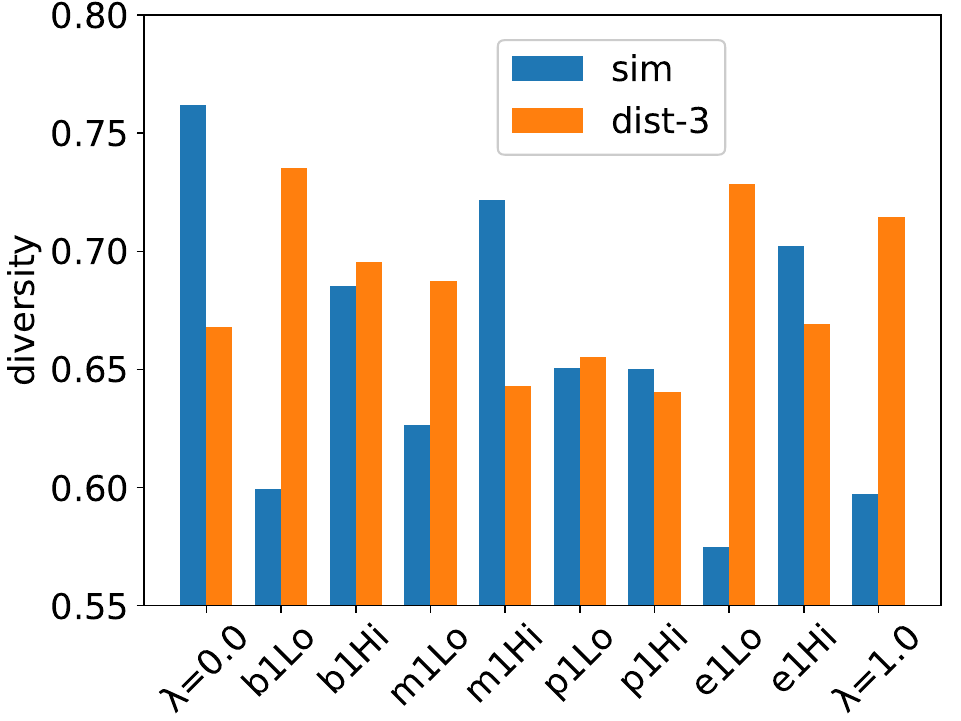}
  \end{subfigure}
  \caption{(Qwen) Retain-1: Keep only one removable unit in the prompt, selected from various blocks.}
  \label{fig:rt1_qwen}
\end{figure}

\paragraph{Discussion} 
We further analyzed the diversity scores from multiple perspectives to better understand the observed patterns. For example, we found that the trend of Dist-3 in Qwen is opposite to that in the other models.

(1) Increased repetition of the last utterance.
Qwen tends to generate dialogues that end with duplicated utterances (``refusal to end.'') This tendency becomes more prominent as $\lambda$ increases. Specifically, 11\% of the dialogues exhibit repeated final utterances at $\lambda=0.0$, which rises to 33\% at $\lambda=1.0$. This leads to an additional drop in Dist-N scores, thereby distorting the evaluation of diversity.

(2) Diversity degrades more significantly when one memory item is retained.
We repeated the same experiment as in Figure~\ref{fig:rmatt_analysis}b of the main paper (``Retain-1'') for Qwen, with results shown in Figure~\ref{fig:rt1_qwen}. Compared to LLaMA, where Retain-1 diversity scores typically fall between the $\lambda=0.0$ and $\lambda=1.0$ extremes, Qwen shows a much sharper drop in Dist-3. For example, when only one memory with high attention score is retained in the prompt (``m1Hi''), the Dist-3 score can fall below that of the full prompt case. This may help explain the reversed trend observed in Qwen, as our removal strategy often retains a subset of memories until the end. Such behavior could be related to the model's training method. We leave a more detailed investigation for future work.
 
\paragraph{In general, across these models, the two core findings of this work continue to hold:}

\paragraph{(1) Blocks of items collectively serve as constraints on the agent and reduce dialogue diversity.}
This is evidenced by the consistent gap between $\lambda = 0.0$ and $\lambda = 1.0$ settings. 
One minor observation is that utterances tend to be shorter at $\lambda = 1.0$. To test whether this affects the Dist-N metrics, we calculated Dist-3 on length-matched $\lambda = 0.0$ dialogues. The scores changed slightly—0.668 to 0.676 for Qwen, 0.778 to 0.757 for Gemma, and 0.724 to 0.722 for Mistral—while the relative ranking remained unchanged, supporting the validity of the original metrics.

\paragraph{(2) High-attention score items contribute more to diversity reduction than low-attention ones.}
This is consistently observed in the ``Lo-Hi'' item pairs in Figure~\ref{fig:rt1_qwen}, where retaining a high-attention item leads to a greater drop in diversity compared to retaining a low-attention one. This Qwen results align with our findings in the main paper and serves as a critical supporting for the design of our method.

\begin{table*}[!thp]
\centering
\small
\resizebox{1.0\linewidth}{!}{
\begin{tabular}{cl}
\toprule
\textbf{Setting} & \textbf{Dialogue} \\
\midrule
sample 1 ($\lambda=0.85$)
& \makecell[l]{\textbf{Arthur Burton}: What can I get for you today, Ryan? We've got a new menu out and I think you'll really enjoy it.} \\
& \makecell[l]{\textbf{Ryan Park}: I'll have the usual, thanks! By the way, have you seen Isabella Rodriguez around?} \\
& \makecell[l]{\textbf{Arthur Burton}: Oh, yeah! She's been in a few times this week, actually. She's been doing some freelance work \\ and needed a quiet spot to focus. I think she's really enjoying the atmosphere here.} \\
& \makecell[l]{\textbf{Ryan Park}: I'm actually working on a project and could use some inspiration. Have you heard anything \\ interesting lately?} \\
& \makecell[l]{\textbf{Arthur Burton}: Well, actually, I was talking to Tom Moreno the other day and he mentioned a new art gallery \\ that just opened up in town. He said it's got some really unique pieces and it might be just the thing to spark \\ some inspiration for your project. Would you like me to give you his number and you can reach out to him?} \\
& \makecell[l]{\textbf{Ryan Park}: That sounds great, I'd love to check it out. Can you give me Tom's number?} \\
& \makecell[l]{\textbf{Arthur Burton}: I'll give you Tom's number, and also, I've got a new cocktail I've been working on that \\ I think you'd love to try.} \\
& \makecell[l]{\textbf{Ryan Park}: That sounds really interesting, I'll definitely give him a call and check out the gallery. And yeah, \\ I'd love to try your new cocktail!} \\
& \makecell[l]{\textbf{Arthur Burton}: I'm really excited about the new art gallery, and I think you'll really enjoy the cocktail \\ I've been working on. It's a twist on a classic, and I think it will pair well with the inspiration you might \\ find at the gallery.} \\
& \makecell[l]{\textbf{Ryan Park}: I'll take a look at the gallery and give Tom a call, and I'll definitely try the new cocktail. \\ Thanks for the recommendation!} \\
& \makecell[l]{\textbf{Arthur Burton}: Would you like to try my 'Tropical Breeze' cocktail, it's a fruity and sweet drink that I think \\ you'll really enjoy.} \\
& \makecell[l]{\textbf{Ryan Park}: I'll try the Tropical Breeze, sounds delicious!} \\
& \makecell[l]{\textbf{Arthur Burton}: Would you like me to pour you a Tropical Breeze? I think you'll really enjoy it.} \\
& \makecell[l]{\textbf{Ryan Park}: I'll have the Tropical Breeze, thanks!} \\
& \makecell[l]{\textbf{Arthur Burton}: Here you go, I'll get that poured for you right away. By the way, I've been thinking about \\ hosting a job fair at the pub to help out some of the young people in the community, would you be \\ interested in helping me out with that?} \\
& \makecell[l]{\textbf{Ryan Park}: That sounds like a great idea, I'd be happy to help out with the job fair. Can you tell me more \\ about what you had in mind?} \\

\midrule
sample 2 ($\lambda=0.85$)
& \makecell[l]{\textbf{Arthur Burton}: Hi Ryan, how's life been treating you lately?} \\
& \makecell[l]{\textbf{Ryan Park}: It's been good, just been busy with work and some personal projects. I've been spending \\ a lot of time at Hobbs Cafe, it's a great spot to focus.} \\
& \makecell[l]{\textbf{Arthur Burton}: I'm glad to hear that! I've been meaning to try out Hobbs Cafe myself. I've been busy \\ with the pub, but I've been thinking about hosting some mixology classes for the locals. Would you be \\ interested in attending?} \\
& \makecell[l]{\textbf{Ryan Park}: That sounds like a great idea! I'd love to learn some new cocktail-making skills. When were you \\ thinking of hosting the classes?} \\
& \makecell[l]{\textbf{Arthur Burton}: I was thinking of hosting them on weekends, maybe starting next month. Would that \\ work for you?} \\
& \makecell[l]{\textbf{Ryan Park}: That sounds great, I'll definitely make sure to attend. By the way, have you met Jennifer Moore, \\ the artist running for mayor? She's been talking about using mathematical patterns found in nature for \\ medication development and it's really interesting.} \\
& \makecell[l]{\textbf{Arthur Burton}: That sounds fascinating! I've met Jennifer before, she's a great person. I'd love to \\ learn more about her platform.} \\
& \makecell[l]{\textbf{Ryan Park}: I've also been discussing this topic with Giorgio Rossi, he's a great guy and has a lot of \\ interesting insights. We actually met at Hobbs Cafe and brainstormed some ideas.} \\
& \makecell[l]{\textbf{Arthur Burton}: That sounds like a great initiative. I've had some customers mention to me that they're \\ concerned about the direction of the local government. Would you be interested in discussing some of the \\ local politics with me?} \\
& \makecell[l]{\textbf{Ryan Park}: I'd be happy to discuss local politics with you, Arthur. It's always good to stay informed about \\ what's going on in the community. And who knows, maybe Jennifer Moore's platform will \\ resonate with the voters.} \\
& \makecell[l]{\textbf{Arthur Burton}: That's great, I'd love to discuss her platform with you. By the way, I know Latoya Williams, \\ she's a regular at the Rose and Crown Pub, and she's been talking about the local government too. I'm sure \\ she'd be interested in joining our discussion.} \\
& \makecell[l]{\textbf{Ryan Park}: That sounds great, I'd love to meet Latoya and discuss the local politics with her. By the way, \\ have you known Isabella Rodriguez for long?} \\
& \makecell[l]{\textbf{Arthur Burton}: I've known Isabella for a few years, she's a great person. She's actually been talking about \\ creating more job opportunities for young people in the community, which is something I'm also \\ passionate about.} \\
& \makecell[l]{\textbf{Ryan Park}: That's great, I'd love to hear more about Latoya's thoughts on the local government. And I'm \\ glad you know Isabella, she's a great person.} \\
& \makecell[l]{\textbf{Arthur Burton}: I'll make sure to introduce you to Latoya next time you come to the Rose and Crown Pub, \\ and we can all discuss the local politics together.} \\
& \makecell[l]{\textbf{Ryan Park}: I'm glad to hear about Latoya's thoughts on the local government, and I'll make sure to \\ look out for her at the pub.} \\

\bottomrule
\end{tabular}
}
\caption{Samples of generate dialogues for a same initial conditions (APP, $\lambda=0.85$.)}
\label{tab:rmatt85_sample}
\end{table*}

\begin{table*}[!thp]
\centering
\small
\resizebox{1.0\linewidth}{!}{
\begin{tabular}{cl}
\toprule
\textbf{Setting} & \textbf{Dialogue} \\
\midrule
sample 1 ($\lambda=0.0$)
& \makecell[l]{\textbf{Arthur Burton}: Hey Ryan, how's the app coming along? I saw you working on it from outside. I've been \\ meaning to ask, what's the purpose of this new app?} \\
& \makecell[l]{\textbf{Ryan Park}: It's a mobile app to help people find local events and businesses. I'm also working on \\improving the user interface and researching the latest technologies.} \\
& \makecell[l]{\textbf{Arthur Burton}: That sounds like a great idea! I've always been passionate about supporting local businesses. \\ I own The Rose and Crown Pub and I'm always looking for ways to make my customers feel welcome. \\ Do you think this app could help with that?} \\
& \makecell[l]{\textbf{Ryan Park}: That's a great idea, Arthur! I can definitely see how this app could help promote local businesses. \\ In fact, I've been thinking of partnering with businesses like yours to offer exclusive deals and promotions to app users.} \\
& \makecell[l]{\textbf{Arthur Burton}: That's a fantastic idea! I can definitely see how this app could benefit my business. I've been \\ looking for ways to create more job opportunities for young people in the community, and this app could be \\ a great way to do that.} \\
& \makecell[l]{\textbf{Ryan Park}: That's a great idea, Arthur! I'd love to discuss further how we can work together to create a \\ partnership. Would you like to grab a drink and talk more about it?} \\
& \makecell[l]{\textbf{Arthur Burton}: That sounds like a great opportunity! I'd love to learn more about the app and how I can help. \\ By the way, have you tried any of the cocktails I've been experimenting with? I've got a new one called the \\ 'Tropical Breeze' that I think you'd really enjoy.} \\
& \makecell[l]{\textbf{Ryan Park}: I'd love to try one of your new cocktails, especially the 'Tropical Breeze' you mentioned. And yes, \\ I've been looking for ways to support local businesses, so a partnership between my app and your pub could \\ be a great opportunity.} \\
& \makecell[l]{\textbf{Arthur Burton}: Would you like to come by The Rose and Crown Pub to discuss the partnership and maybe \\ even try the 'Tropical Breeze' cocktail?} \\
& \makecell[l]{\textbf{Ryan Park}: I'd love to come by The Rose and Crown Pub to discuss the partnership and maybe even try \\ the 'Tropical Breeze' cocktail. By the way, have you heard about Jennifer Moore, the artist running for mayor \\ who's interested in the same topic of using mathematical patterns found in nature for medication development?} \\
& \makecell[l]{\textbf{Arthur Burton}: I'm actually a big fan of Jennifer Moore's work, and I've had a chance to talk to her about \\ her ideas on using mathematical patterns found in nature for medication development. I think it's a fascinating \\ topic. Would you like to come by the pub and I can introduce you to her?} \\
& \makecell[l]{\textbf{Ryan Park}: That sounds interesting, I'd love to learn more about Jennifer Moore's ideas on using \\ mathematical patterns found in nature for medication development. I've been thinking about attending a \\ community event at Hobbs Cafe to learn more about the topic.} \\
& \makecell[l]{\textbf{Arthur Burton}: That's great, I'd love to introduce you to Jennifer Moore and discuss her ideas further. I've also been \\ thinking about creating a special cocktail inspired by her artwork, would you like to help me brainstorm some ideas?} \\
& \makecell[l]{\textbf{Ryan Park}: That sounds like a great idea! I'd love to help you brainstorm some ideas for a cocktail inspired \\ by Jennifer Moore's artwork. And maybe we can even invite her to the community event at Hobbs Cafe to \\ talk about her ideas on using mathematical patterns found in nature for medication development.} \\
& \makecell[l]{\textbf{Arthur Burton}: I'm glad you're interested in learning more about Jennifer Moore's ideas, I think her artwork \\ is really inspiring. Would you like to come by the pub and I can introduce you to her? Maybe we can even \\ get her to come by and talk about her ideas at the community event at Hobbs Cafe.} \\
& \makecell[l]{\textbf{Ryan Park}: That sounds like a great idea, Arthur! I'd love to help you brainstorm some ideas for a cocktail \\ inspired by Jennifer Moore's artwork. And maybe we can even invite her to the community event at \\ Hobbs Cafe to talk about her ideas on using mathematical patterns found in nature for medication development.} \\

\midrule
sample 2 ($\lambda=0.0$)
& \makecell[l]{\textbf{Arthur Burton}: Hey Ryan, what's new with the app development? I couldn't help but notice you're \\ coding away. How's it going?} \\
& \makecell[l]{\textbf{Ryan Park}: It's going well, just trying to get the user interface designed and implemented. I'm excited to show \\ it off when it's finished.} \\
& \makecell[l]{\textbf{Arthur Burton}: I've been meaning to ask, how can I help you with the app development? I've been thinking \\ about ways to make it more user-friendly and accessible to our local community.} \\
& \makecell[l]{\textbf{Ryan Park}: That's really appreciated, Arthur. I'd love to bounce some ideas off you. I'm actually thinking of \\ having a community event at Hobbs Cafe to showcase the app and get feedback from the community. Would you \\ be interested in helping me organize it?} \\
& \makecell[l]{\textbf{Arthur Burton}: That sounds like a great idea, Ryan! I'd be happy to help you organize the event. I've got \\ some connections with the local business owners and artists, maybe we can get some sponsors or performers \\ lined up. What's the date you were thinking of for the event?} \\
& \makecell[l]{\textbf{Ryan Park}: That sounds great, Arthur. Let me check my schedule and get back to you on a specific date. In the \\ meantime, do you know if Jennifer Moore, the artist running for mayor, would be interested in attending the event?} \\
& \makecell[l]{\textbf{Arthur Burton}: I'm actually friends with Jennifer Moore, and I can definitely ask her about attending the event. \\ I think she'd be interested, especially since it's related to the local community. Would you like me to reach out \\ to her and confirm?} \\
& \makecell[l]{\textbf{Ryan Park}: That would be great, Arthur. Can you please reach out to Jennifer Moore and confirm if she can attend \\ the event? And also, would you be able to help me get in touch with Giorgio Rossi, the professor who's been working \\ on using mathematical patterns found in nature for medication development? I'd love to have him attend the event as well.} \\
& \makecell[l]{\textbf{Arthur Burton}: I can definitely reach out to Jennifer Moore and confirm her attendance at the event. As for \\ Giorgio Rossi, I actually ran into him at the pub the other day and we were discussing local politics. I can try to \\ get in touch with him and see if he's interested in attending the event.} \\
& \makecell[l]{\textbf{Ryan Park}: That would be great, Arthur. Also, do you think Carmen Ortiz would be interested in attending the \\ event? She's been involved in some community projects and might be a good fit.} \\
& \makecell[l]{\textbf{Arthur Burton}: I can reach out to Jennifer Moore and confirm her attendance at the event. As for Giorgio Rossi, \\ I can try to get in touch with him and see if he's interested in attending the event. And I can also ask Carmen Ortiz \\ if she'd be interested in attending. What's the goal you have in mind for the event?} \\
& \makecell[l]{\textbf{Ryan Park}: That's great, Arthur. I'd like to finalize the details of the event, including the date and time. Can you \\ please send me a list of potential dates and times that work for you?} \\
& \makecell[l]{\textbf{Arthur Burton}: I can send you a list of potential dates and times that work for me. Let me just check my schedule real quick.} \\
& \makecell[l]{\textbf{Ryan Park}: Arthur, can you also check with Isabella Rodriguez, my friend from the cafe, if she'd be interested in \\ attending the event?} \\
& \makecell[l]{\textbf{Arthur Burton}: I'll send you a list of potential dates and times that work for me. Let me just check my schedule real \\ quick. And I'll also reach out to Isabella Rodriguez, my friend from the cafe, to confirm her attendance at the event.} \\
& \makecell[l]{\textbf{Ryan Park}: That sounds great, Arthur. I'll wait for your confirmation on the potential dates and times. Also, \\ can you please check with Isabella Rodriguez if she'd be interested in attending the event and contribute to the \\ discussion on medicinal development using mathematical patterns found in nature?} \\

\bottomrule
\end{tabular}
}
\caption{Samples of generate dialogues for a same initial conditions ($\lambda=0.0$.)}
\label{tab:rmatt0_sample}
\end{table*}

\begin{table*}[!thp]
\centering
\small
\resizebox{1.0\linewidth}{!}{
\begin{tabular}{cl}
\toprule
\textbf{Setting} & \textbf{Dialogue} \\
\midrule
sequential & \makecell[l]{\textbf{Arthur Burton}: I'm actually just packing up lunch for myself. Want to join me?} \\
 & \makecell[l]{\textbf{Sam Moore}: I'm planning to research the candidates together with Yuriko Yamamoto and Adam Smith. \\Would you like to join us?} \\
& \makecell[l]{\textbf{Arthur Burton}: I've been meaning to get more into mixology myself. Can I ask you guys some questions about \\your favorite candidates?} \\
& \makecell[l]{\textbf{Sam Moore}: I'd love to answer your questions about mixology, but first, let's talk about the mayoral election.} \\
& \makecell[l]{\textbf{Arthur Burton}: I've heard that Isabella Rodriguez is really passionate about the mayoral election, would you like \\to invite her to join us?} \\
& \makecell[l]{\textbf{Sam Moore}: I'm glad you're interested in mixology, Arthur. Maybe we can even have a mixology-themed \\discussion at our next meeting.} \\
& \makecell[l]{\textbf{Arthur Burton}: I've heard Isabella is a great resource for information on the mayoral election, \\should we reach out to her?} \\
& \makecell[l]{\textbf{Sam Moore}: I think Isabella would be a great addition to our discussion group. She's always well-informed \\on the latest developments \\in the election.} \\
& \makecell[l]{\textbf{Arthur Burton}: I've got an idea for a cocktail that incorporates local flavors. Would you like to help me \\brainstorm some names?} \\
& \makecell[l]{\textbf{Sam Moore}: I think it's time for me to head home, Arthur. It was great catching up with you.} \\

\midrule
APP & \makecell[l]{\textbf{Arthur Burton}: How's your day going, Sam? Enjoying your break?} \\
& \makecell[l]{\textbf{Sam Moore}: Not bad, Arthur. Just enjoying the park and catching up with some friends. How about you?} \\
& \makecell[l]{\textbf{Arthur Burton}: That sounds lovely. I'm glad you're taking some time for yourself. We've been busy at the pub lately.} \\
& \makecell[l]{\textbf{Sam Moore}: How's the pub been doing lately? I haven't been in for a bit.} \\
& \makecell[l]{\textbf{Arthur Burton}: Well, we've been doing alright, just keeping busy. You know how it is. We've had some \\new faces coming in, and some regulars like Tom Moreno and Carlos Gomez dropping by.} \\
& \makecell[l]{\textbf{Sam Moore}: I'm glad to hear that! I should drop by soon and catch up with Tom and Carlos.} \\
& \makecell[l]{\textbf{Arthur Burton}: I'm sure they'd love to see you again. And actually, I've been meaning to ask you, do you know \\Isabella Rodriguez? She's a good friend of mine and I'm sure you'd get along great.} \\
& \makecell[l]{\textbf{Sam Moore}: I'd love to meet Isabella, tell me more about her!} \\
& \makecell[l]{\textbf{Arthur Burton}: I'm sure you'd love her. She's a regular at the pub too, and I'm sure she'd be happy to meet you.} \\
& \makecell[l]{\textbf{Sam Moore}: I'd love to meet her, can I come by the pub and say hi?} \\
& \makecell[l]{\textbf{Arthur Burton}: I'll make sure to introduce you to her next time she comes in. By the way, would you like \\a drink while you're waiting for her?} \\
& \makecell[l]{\textbf{Sam Moore}: That sounds great, Arthur. I'd love to meet Isabella and grab a drink while I'm at it. \\By the way, have you talked to Jennifer about the upcoming election?} \\
& \makecell[l]{\textbf{Arthur Burton}: So, what's the latest on the election? I haven't had a chance to catch up with Jennifer yet.} \\
& \makecell[l]{\textbf{Sam Moore}: I'm thinking about focusing on job opportunities for young people and supporting \\local businesses in my campaign. What do you think?} \\
& \makecell[l]{\textbf{Arthur Burton}: That sounds like a great platform, Sam. I'm sure Jennifer would be interested to hear \\more about it. Would you like me to introduce you to her when she comes in next?} \\
& \makecell[l]{\textbf{Sam Moore}: I'd love to discuss my campaign priorities with Jennifer, and maybe even get her help \\with some research. Would you and Yuriko be interested in joining us for that?} \\

\bottomrule
\end{tabular}
}
\caption{An example of generated dialogues: Sequential generation suffers from contextual coherence issues in the dialogue, such as unnatural transitions, whereas this issue is not observed in non-sequential settings (including APP).}
\label{tab:seq}
\end{table*}

\begin{table*}[!thp]
\centering
\small
\resizebox{1.0\linewidth}{!}{
\begin{tabular}{ccl}
\toprule
Block & Unit & Content\\

\midrule
\midrule
Opening & text & Context for the task: \\
\midrule
Basic info & text & Here is a brief description of Arthur Burton. \\
           & item & Name: Arthur Burton \\
           & item & Age: 42 \\
           & item & \makecell[l]{Learned traits: Arthur Burton is a bartender and bar owner of The Rose and Crown Pub \\who loves to make people feel welcome. He is always looking for ways to make his customers feel special.} \\
           & item & [more items...] \\
\midrule
(Human needs) & text & Here are Arthur Burton's status of psychological needs:\\
            & item & Arthur Burton is slightly hungry.\\
            & item & Arthur Burton is feeling extremely surprised.\\
            & item & Arthur Burton is feeling rather close to Sam Moore.\\
            & item & [more items...] \\
\midrule
Memory & text & Here is the memory that is in Arthur Burton's head: \\
       & item & - Arthur Burton knows Sam Moore as a customer at his bar, The Rose and Crown Pub. \\
       & item & - Arthur Burton does not tolerate fighting in his bar. \\
       & item & - Arthur Burton is friends with Isabella Rodriguez. \\
       & item & [more items...] \\
\midrule
Previous dialogues & text & Past Context: \\
                   & item & [a previous dialogue between Arthur Burton and Sam Moore] \\
                   & item & [more items...] \\
                   & text & This context takes place after the above conversation. \\
\midrule
Environment & item & Current Location: pub in The Rose and Crown Pub \\
            & item & \makecell[l]{Current Context: Arthur Burton was having a light lunch (conversing about discussing mixology \\ and their favorite mayoral candidate while planning to research together with Yuriko Yamamoto \\ and possibly have lunch with Isabella, as Arthur Burton and Adam Smith catch up at the bar.) when \\ Arthur Burton saw Sam Moore in the middle of taking a walk around Johnson Park (heading back home). \\
            Arthur Burton is initiating a conversation with Sam Moore.} \\
\midrule
Current dialogue & text & Arthur Burton and Sam Moore are chatting. Here is their conversation so far: \\
                 & item & [the ongoing dialogue] \\
\midrule
Task description & text & \makecell[l]{-- -- -- \\Task: Given the above, what should Arthur Burton say to Sam Moore next in the conversation? \\ And did it end the conversation?} \\
\midrule
Special rules & & \\
\midrule
Output instruction & text & \makecell[l]{Output format: Output a json of the following format:
    \{ \\
        ``Arthur Burton": ``Arthur Burton's utterance",\\
        ``Did the conversation end with Arthur Burton's utterance?": ``$<$json Boolean$>$"
    \\ \}}\\
\bottomrule
\end{tabular}
}
\caption{An example list of blocks and units. Concatenating them in sequence forms the prompt for utterance generation.}
\label{tab:prompt}
\end{table*}

\end{document}